\pdfoutput=1
\documentclass[10pt,twocolumn,letterpaper]{article}
\usepackage[T1]{fontenc}
\usepackage[utf8]{inputenc}
\usepackage[margin=0.72in,columnsep=0.25in]{geometry}
\usepackage{lmodern,microtype}
\usepackage{amsmath,amssymb,amsfonts}
\usepackage{amsthm}                 %
\usepackage{booktabs,tabularx,array}
\usepackage{graphicx,xcolor,soul}
\usepackage{tikz}
\usetikzlibrary{positioning,fit,backgrounds,arrows.meta,calc}
\usepackage{enumitem}
\usepackage{caption,sectsty}
\allsectionsfont{\raggedright}
\usepackage{algorithm}              %
\usepackage{algpseudocode}
\usepackage[caption=false,font=normalsize,labelfont=sf,textfont=sf]{subfig}
\usepackage{textcomp}
\usepackage{stfloats}
\usepackage{verbatim}
\usepackage{cite}
\usepackage{xspace}
\usepackage[hidelinks]{hyperref}    %
\usepackage{xurl}

\theoremstyle{plain}
\newtheorem{theorem}{Theorem}
\newtheorem{lemma}{Lemma}
\newtheorem{proposition}{Proposition}
\newtheorem{corollary}{Corollary}
\theoremstyle{definition}

\theoremstyle{remark}

\providecommand{\appendices}{\appendix}

\newcolumntype{C}{>{\centering\arraybackslash}X}
\newcolumntype{Y}{>{\raggedright\arraybackslash}X}

\numberwithin{equation}{section}

\setlist{nosep,leftmargin=*}
\newif\ifshowreview
\showreviewfalse
\newcounter{rvnote}

\newcommand{\yz}[2][blue]{\ifshowreview\textcolor{#1}{\textbf{[YZ: #2]}}\fi}

\title{\vspace{-1.5em}\textbf{Class-aware Latent Attention Replay (CLAR): Class
Incremental Hazard Diagnosis for Resilient AI in
the Manufacturing Industrial Internet}}
\author{Avranshu Mahanty\textsuperscript{a}, Ran Jin\textsuperscript{b},  Yingyan Zeng\textsuperscript{a,*}\\[4pt]
\small\textsuperscript{a}Department of Mechanical, Materials, and Industrial Engineering, University of Cincinnati, Cincinnati, OH\\
\small\textsuperscript{b}Department of Industrial and Systems Engineering, Virginia Tech, Blacksburg, VA\\
}
\date{}

\begin{document}
\maketitle

\begin{abstract}
Artificial Intelligence (AI) systems provide critical computation services in the Manufacturing Industrial Internet (MII), and their resilience depends on timely diagnosis of hazards across the data, AI pipeline, and cyber-physical layers. During operation, new hazard types keep emerging with only a few samples each, hazards co-occur, and the diagnosis model runs on memory-constrained computation nodes. 
Replay-based class-incremental learning (CIL) can learn new hazard classes while revisiting a small memory of past samples, but existing methods assume single-label, balanced streams, store high-dimensional raw samples, and select them independently of the classifier. 
We propose Class-aware Latent Attention Replay (CLAR), which combines a dual buffer that stores mostly compact latent representations with a few raw samples, an Attention Leverage Score (ALS) that upper-bounds the classifier logit and guides class-balanced reservoir sampling, and an exponential-moving-average teacher that slows latent-space drift. For an idealized buffer model, we prove that the buffer is class-balanced and follows an ALS-weighted class distribution, and that below a critical per-class capacity, ALS-guided replay has a lower replay-gradient error  than class-balanced uniform replay. In the case study, On hazard diagnosis data from an MII testbed, CLAR achieves the highest mean average weighted F1 score among the evaluated replay methods with the same buffer size.
\end{abstract}

\section{Introduction}\label{sec:intro}

The Manufacturing Industrial Internet (MII) connects manufacturing equipment, physical processes, systems, and networks through ubiquitous sensors, actuators, and computing units \cite{chen2018predictive, zeng2025fair}.
The data collected via the MII enable AI models that deliver computation services such as quality modeling, process variation analysis, and fault diagnosis \cite{arinez2020artificial}.
In this context, an \textit{AI system} consists of three interconnected layers \cite{chen2020adapipe, zeng2025fair}: (i)~the data layer, i.e., the manufacturing data collected via the MII; (ii)~the AI pipeline layer, where the data are processed by deployed AI pipelines, such as data augmentation, standardization, and deep neural network (DNN) classifiers \cite{shojaee2021deep}; and (iii)~the cyber-physical layer, i.e., the Fog--Cloud computing infrastructure that supports communication and computation.
Each execution of an AI pipeline on a batch of data at a computation node constitutes a \textit{computation task}, e.g., predicting the quality of the products produced by one machine.

Manufacturing decisions, such as accepting a printed part or adjusting a process setting, rely on the outputs of these computation tasks, so the AI system has to deliver reliable computation services continuously throughout production.
Hazards in any of the three layers threaten this reliability.
Sensor contamination and low signal-to-noise ratio corrupt the data layer, model singularity destabilizes the AI pipeline layer, and fog node or communication failures disrupt the cyber-physical layer \cite{zeng2025fair}.
The resulting performance degradation leads to incorrect quality predictions, faulty process decisions, and economic losses.
\textit{AI resilience} refers to the ability of the AI system to withstand, respond to, and recover from such hazards \cite{poulin2021infrastructure, zeng2025fair}.
Recovery requires a mitigation action that matches the root cause, so accurate root-cause diagnosis is the prerequisite for resilience.
Diagnosis can use information that the MII already collects after each computation task: the supervised learning performance of the deployed pipelines and the runtime metrics of the computation node, such as CPU utilization and memory consumption \cite{zeng2025fair}.
Hazard diagnosis therefore becomes a multimodal, multi-label classification problem over these two modalities.

Diagnosis models of this kind are trained offline on a fixed set of hazard types.
In operation, the MII keeps changing: new machines, process settings, and software updates introduce hazard types that were absent from the training data, and known hazard types reappear later.
A resilient AI system thus needs a diagnosis model that keeps learning from the stream of computation tasks, which raises five challenges.
(i)~New hazard classes emerge over time and previously seen classes recur, and retraining from scratch on all historical data is computationally prohibitive.
(ii)~Normal operation dominates the stream, so each hazard class has only a few samples.
(iii)~Several hazards can affect the same computation task, making the problem multi-label.
(iv)~The runtime metrics are high-dimensional multivariate time series, which, together with the tabular performance metrics, form multimodal samples.
(v)~The diagnosis model runs on resource-constrained computation nodes, which strictly limits its memory.

Class-Incremental Learning (CIL) addresses the first challenge by updating a model with new classes as they appear \cite{de2021continual, masana2022class}.
Its central difficulty is \textit{catastrophic forgetting}: training on new classes overwrites the knowledge of earlier classes and degrades their accuracy \cite{mccloskey1989catastrophic}.
Replay-based methods counter forgetting by storing a small memory of past samples and revisiting them while learning new classes \cite{chaudhry2019continual, rebuffi2017icarl}.
Three limitations restrict their use for hazard diagnosis in the MII.
First, most methods assume single-label streams with balanced and sufficient samples per class, while hazard data are multi-label, highly imbalanced, and scarce.
Second, they store past samples in raw form; a raw sample in our case study contains about $2.5\times10^{6}$ values, so a fixed memory budget holds only a few exemplars per hazard class.
Third, they select samples to store by random, class-count, gradient, or class-mean criteria, which are independent of the class-specific parameters that the classifier uses to make its decisions.

To address these limitations, we propose Class-aware Latent Attention Replay (CLAR), a class-incremental hazard diagnosis framework for AI systems in the MII.
The key idea of CLAR is to keep the memory of past hazards predominantly in the compact latent space of the diagnosis model and to let the class-specific attention parameters of the classifier decide which samples are worth remembering.
The contributions of this work are as follows.
\begin{itemize}
    \item We formulate class-incremental hazard diagnosis for AI systems in the MII as a multi-label, multimodal CIL problem over a sequential stream of tasks with recurring classes, severe class imbalance, and a strict memory budget.
    \item We propose a latent-dominant dual replay buffer that stores many compact latent representations and a few raw samples per hazard class, which multiplies the number of buffer slots per class within the same memory budget.
    \item We propose the Attention Leverage Score (ALS), an attention-induced Mahalanobis-type norm that upper-bounds the magnitude of the classifier logit, and use it to tilt class-balanced reservoir sampling toward samples that strongly drive the classifier under imbalanced, multi-label streams.
    \item We show that ALS-guided sampling keeps a class-balanced buffer whose content follows an ALS-weighted class distribution, decompose its replay-gradient error into sampling variance and selection bias, and characterize, for an idealized model, the per-class capacities below which it has a lower replay-gradient error and a smaller upper bound on the expected squared margin deviation than class-balanced uniform replay.
    \item We evaluate CLAR on hazard data from an MII testbed, where it achieves the highest mean average weighted F1 score among the evaluated methods with a smaller configured buffer.
\end{itemize}

The remainder of this paper is organized as follows.
Section~\ref{sec:related} reviews CIL, replay-based CIL, and few-shot CIL.
Section~\ref{sec:method} formulates the problem and presents CLAR.
Section~\ref{sec:theory} analyzes ALS-guided sampling theoretically.
Section~\ref{sec:exp} reports the case study, and Section~\ref{sec:conclusion} concludes the paper.

\begin{figure*}[t]
    \centering
    \includegraphics[width=0.9\linewidth]{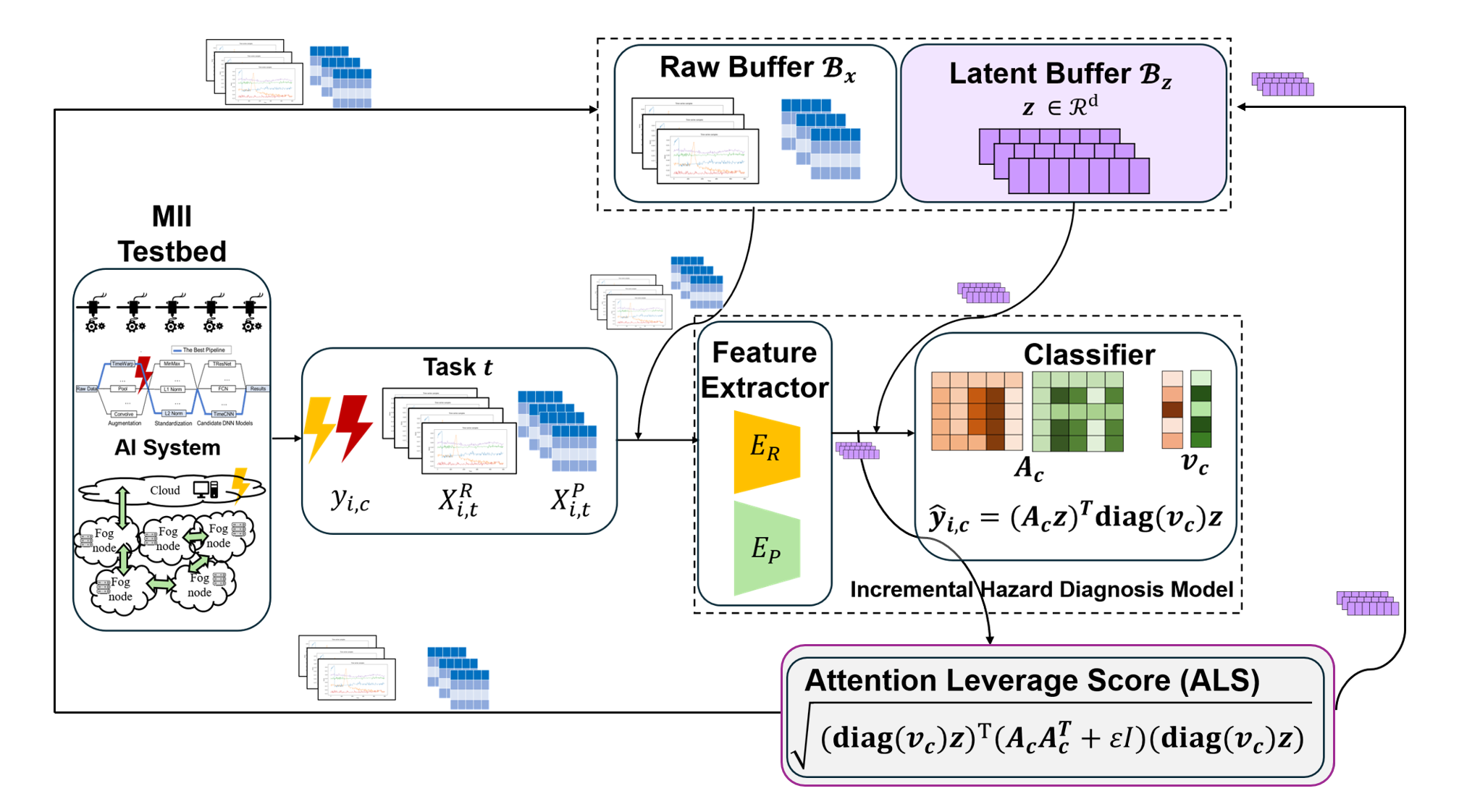}
    \caption{Overview of the incremental diagnosis model, MII testbed, and the CLAR framework.}
    \label{fig:wide_figure}
\end{figure*}
\section{Related Work}\label{sec:related}

\subsection{Class-Incremental Learning}
Continual learning trains a model on a sequence of tasks, where each task provides a new set of training data and the data of earlier tasks become unavailable \cite{de2021continual}.
CIL is the setting of continual learning in which new tasks introduce new classes and the model has to classify a test sample among all classes seen so far, without being told which task the sample comes from \cite{van2022three, masana2022class}.
This setting is the most challenging one, because the model has to separate classes that were never trained together, in addition to learning each task.
It matches applications where new categories emerge during deployment, such as new defect types in visual inspection or new fault types in machinery diagnosis \cite{yan2024few}.

Two related phenomena make CIL difficult.
Catastrophic forgetting occurs when gradient updates for new classes change the parameters that encode earlier classes, so the accuracy on earlier classes drops sharply \cite{mccloskey1989catastrophic}.
Avoiding it requires stability, whereas learning new classes requires plasticity, and the trade-off between the two is known as the stability--plasticity dilemma \cite{mermillod2013stability}.
At the output layer, forgetting appears as task-recency bias: since the training data of each task are dominated by the newest classes, the classifier assigns higher scores to recent classes and misclassifies samples of earlier classes as recent ones \cite{wu2019large, hou2019learning}.
Practical streams add further difficulties.
In online CIL, each sample is observed once in a single pass \cite{mai2022online}.
In blurry CIL, classes recur across tasks \cite{bang2021rainbow}; in manufacturing, for example, a sensor fault can reappear after maintenance or when a process setting is reused.
In multi-label CIL, samples carry several labels and the label distribution is long-tailed \cite{kim2020imbalanced}.

CIL methods fall into three families \cite{de2021continual}.
Architecture-based methods expand the network for each new task and freeze or isolate the parameters of earlier tasks \cite{rusu2016progressive, yoon2017lifelong}.
Since the parameters that encode earlier classes are left untouched by later updates, forgetting is avoided by construction.
The model size, however, grows with the number of tasks, which becomes prohibitive on memory-constrained computation nodes.
Regularization-based methods add a penalty to the training loss that discourages changes to the knowledge of earlier tasks.
The penalty can act on parameters, e.g., a squared distance between the current and previous parameters weighted by their estimated importance for earlier tasks \cite{kirkpatrick2017overcoming, zenke2017continual}, or on outputs, e.g., a distillation loss between the predictions of the current model and of a frozen copy, acting as a teacher, on the new data \cite{li2017learning, hinton2015distilling}.
Parameter penalties accumulate over tasks and progressively restrict the capacity to learn new classes, and output penalties are only as informative as the new data are about earlier classes, so their effect weakens under large distribution shifts between tasks.
Both variants train on new-class data alone, which leaves task-recency bias unaddressed \cite{wu2019large, masana2022class}.
Replay-based methods, the third family, address this gap by revisiting samples of earlier classes; as they are the basis of this work, we review them separately.

\subsection{Replay-Based Class-Incremental Learning}
Replay-based methods maintain a \textit{memory buffer}, a fixed-capacity store of past samples.
When the model learns a new task, samples drawn from the buffer are trained together with the new data, so the training objective approximates the objective over all classes seen so far and the classifier keeps observing earlier classes \cite{chaudhry2019continual, rebuffi2017icarl}.
The design of a replay method answers two questions: which samples to store (memory update) and which stored samples to replay (memory retrieval) \cite{mai2022online}.

For memory update, Experience Replay (ER) \cite{chaudhry2019continual} uses reservoir sampling \cite{vitter1985random}, which keeps a uniform random subset of the stream by admitting the $n$-th sample with probability proportional to $1/n$.
Under an imbalanced stream, a uniform subset inherits the imbalance and minority classes nearly vanish from the buffer.
Class-Balancing Reservoir Sampling (CBRS) \cite{chrysakis2020online} therefore applies reservoir sampling within each class and reserves equal space for every class, and partitioning reservoir sampling extends this idea to multi-label streams \cite{kim2020imbalanced}.
These strategies choose samples at random within a class.
Informativeness-driven strategies select samples by gradient diversity \cite{aljundi2019gradient}, by diversity and uncertainty \cite{bang2021rainbow}, or by proximity to the class mean in the feature space, as in the herding selection \cite{welling2009herding} of iCaRL \cite{rebuffi2017icarl}.
Feature-space criteria become unreliable as the feature extractor is fine-tuned on new classes, since the embedding of stored samples drifts.
For memory retrieval, Maximally Interfered Retrieval (MIR) \cite{aljundi2019online} replays the stored samples whose loss would increase most under the upcoming update, and ASER \cite{shim2021online} ranks stored samples by adversarial Shapley values.

Most replay methods store raw inputs.
Latent replay stores intermediate representations of a network, which are much smaller and therefore fit more exemplars into the same memory \cite{pellegrini2020latent, hayes2020remind}.
To keep the stored representations valid, these methods freeze or slowly update the layers below the replay layer, which limits the adaptation of the feature extractor to new classes.
In summary, existing replay methods select samples independently of the class-specific parameters of the classifier, mostly assume single-label streams, and trade memory efficiency against feature adaptation.

\subsection{Few-Shot Class-Incremental Learning}
When new classes arrive with only a handful of labeled samples, CIL becomes Few-Shot Class-Incremental Learning (FS-CIL) \cite{tian2024survey}.
Industrial faults are a typical example, since hazards and defects occur rarely compared with normal operation.
Prototype-based approaches compute a center, or prototype, for each class from its few samples and classify new samples by their distance to the prototypes \cite{snell2017prototypical, rusu2018meta}.
Feature-space-based approaches reserve room in the embedding space for future classes with virtual prototypes \cite{zhou2022few}, which provides forward compatibility, i.e., new class prototypes can be added with little disturbance to the existing representations.
For industrial fault diagnosis, Forward-Back Compatible Representation (FBCR) \cite{yan2024few} combines such feature-space reservation with replay.
All these approaches depend on a stable feature extractor: once the representation space drifts during incremental updates, the prototypes and reserved regions become outdated and the diagnosis accuracy degrades.

\section{Methodology}\label{sec:method}

This section formalizes class-incremental hazard diagnosis for an AI system in the MII (Section~\ref{sec:setting}), states the key idea of CLAR (Section~\ref{sec:overview}), describes the diagnosis model that CLAR updates over time (Section~\ref{sec:model}), and presents the components of CLAR and its training procedure (Sections~\ref{sec:memory}--\ref{sec:procedure}).

\subsection{Problem Setting}\label{sec:setting}

\textit{Samples and labels.}
For the AI system defined in Section~\ref{sec:intro}, each completed computation task yields one diagnosis sample with two modalities \cite{zeng2025fair}.
The supervised learning performance of the deployed pipelines on the processed batch, such as accuracy, precision, and F1 score, is reported to the Cloud as a tabular vector $X_{i,t}^{(P)}\in\mathbb{R}^{P}$.
The runtime metrics recorded by the computation node during execution, such as CPU utilization, memory consumption, and data transmission volume, form a multivariate time series $X_{i,t}^{(R)}\in\mathbb{R}^{N\times R\times Q}$ with $N$ time stamps, $R$ runtime metrics, and $Q$ channels.
The raw sample with index $i$ in task $t$ is
\begin{equation}
X_{i,t}=\{X_{i,t}^{(P)},X_{i,t}^{(R)}\}.
\end{equation}
These modalities are complementary: hazards in the data and AI pipeline layers mainly alter the supervised learning performance, whereas resource-related hazards mainly alter the runtime behavior.
Each hazard class $c\in\{1,\dots,C\}$ corresponds to one root cause factor of AI system failure \cite{zeng2025fair}, such as sensor contamination or pipeline singularity.
Since hazards can co-occur, the label of $X_{i,t}$ is a binary vector $y_i=(y_{i,1},\dots,y_{i,C})\in\{0,1\}^{C}$, where $y_{i,c}=1$ indicates the presence of hazard class $c$ and $y_i=\mathbf{0}$ indicates normal operation.

\textit{Tasks and incremental classes.}
The samples arrive as a sequential data stream $\mathcal{D}=\{\mathcal{D}_0,\mathcal{D}_1,\dots,\mathcal{D}_T\}$, where each task $\mathcal{D}_t$ collects the samples of one operational phase, $\mathcal{D}_0$ is the base task, and $\mathcal{D}_t$ with $t>0$ is an incremental task.
Let $\mathcal{Y}_t=\{c: y_{i,c}=1 \text{ for some } X_{i,t}\in\mathcal{D}_t\}$ be the hazard classes present in task $t$, $\mathcal{C}_t=\bigcup_{k=0}^{t}\mathcal{Y}_k$ the classes seen up to task $t$, and $\mathcal{C}_{\mathrm{new}}=\mathcal{C}_t\setminus\mathcal{C}_{t-1}$ the new classes introduced by task $t$.
A task can introduce zero or more new classes, and classes can recur, so in general $\mathcal{Y}_k\cap\mathcal{Y}_t\neq\emptyset$ for $k\neq t$.
Each task is partitioned into disjoint training and test sets, $\mathcal{D}_t=\mathcal{D}_t^{\text{train}}\cup\mathcal{D}_t^{\text{test}}$ with $\mathcal{D}_t^{\text{train}}\cap\mathcal{D}_t^{\text{test}}=\emptyset$.
While learning task $t$, the model can access $\mathcal{D}_t^{\text{train}}$ and a memory of fixed capacity, and the training data of earlier tasks are unavailable.
The objective is that the model obtained after task $t$ diagnoses all classes in $\mathcal{C}_t$ accurately on the cumulative test set $\mathcal{D}_t^{\text{eval}}=\bigcup_{k=0}^{t}\mathcal{D}_k^{\text{test}}$.
The model must therefore acquire the classes in $\mathcal{C}_{\mathrm{new}}$ (plasticity) while retaining the classes in $\mathcal{C}_{t-1}$ (stability).

\textit{Assumptions.}
Following the challenges in Section~\ref{sec:intro}, we assume that (i)~the stream is heavily imbalanced toward normal operation, with only a few samples per hazard class; (ii)~samples are high-dimensional and multimodal; (iii)~hazards may co-occur, i.e., the problem is multi-label; (iv)~tasks can introduce zero or more new classes and earlier classes can recur; and (v)~the memory available for storing past samples is strictly limited.

\subsection{Overview of CLAR}\label{sec:overview}

\begin{figure*}[t]
\centering
\usetikzlibrary{positioning,fit,backgrounds,arrows.meta,calc}
\resizebox{\textwidth}{!}{%
\begin{tikzpicture}[
  font=\sffamily\small,
  >={Stealth[length=2.2mm]},
  box/.style={draw, rounded corners=2pt, align=center, minimum height=8mm, inner sep=3pt, fill=white},
  layer/.style={box, minimum width=34mm, fill=gray!8},
  data/.style={box, minimum width=30mm, fill=blue!5},
  enc/.style={box, minimum width=22mm, fill=blue!10},
  prop/.style={box, fill=orange!12, draw=orange!70!black},
  lbl/.style={font=\sffamily\scriptsize, align=center},
  replay/.style={->, dashed, thick, orange!70!black},
  flow/.style={->, thick}
]
\node[layer] (cp) at (0, 1.1) {Cyber-physical layer\\[-1pt]\scriptsize (Fog nodes, Cloud)};
\node[layer] (pl) at (0, 0)   {AI pipeline layer\\[-1pt]\scriptsize (augment., std., DNN)};
\node[layer] (dl) at (0,-1.1) {Data layer\\[-1pt]\scriptsize (MII sensor data)};
\begin{scope}[on background layer]
  \node[draw, rounded corners=3pt, fit=(cp)(pl)(dl), inner sep=6pt, label={[font=\sffamily\small\bfseries]above:AI system in MII \cite{zeng2025fair}}] (sys) {};
\end{scope}

\node[data] (xp) at (5.1, 0.9) {$X_{i,t}^{(P)}$\\[-1pt]\scriptsize supervised learning performance};
\node[data] (xr) at (5.1,-0.9) {$X_{i,t}^{(R)}$\\[-1pt]\scriptsize runtime metrics (time series)};
\draw[flow] (sys.east |- xp) -- (xp.west);
\draw[flow] (sys.east |- xr) -- (xr.west);
\node[lbl] at (2.85,0) {after\\computation\\task $i$};

\node[enc] (ep) at (8.6, 0.9) {$\mathrm{Encoder}_P$\\[-1pt]\scriptsize (VAE)};
\node[enc] (er) at (8.6,-0.9) {$\mathrm{Encoder}_R$\\[-1pt]\scriptsize (LSTM-VAE)};
\node[box, minimum width=17mm, fill=blue!5] (z) at (11.1,0) {$z_i\in\mathbb{R}^{d}$\\[-1pt]\scriptsize latent repr.};
\node[enc, minimum width=30mm] (clf) at (14.2,0) {Attention-based classifier\\[-1pt]$s_c(z_i)$, $\{(\mathbf{A}_c,\mathbf{v}_c)\}$};
\draw[flow] (xp) -- (ep);
\draw[flow] (xr) -- (er);
\draw[flow] (ep.east) -- (z.north west);
\draw[flow] (er.east) -- (z.south west);
\draw[flow] (z) -- (clf);
\begin{scope}[on background layer]
  \node[draw, rounded corners=3pt, fill=blue!2, fit=(ep)(er)(z)(clf), inner sep=6pt,
        label={[font=\sffamily\small\bfseries]above:Incremental hazard diagnosis model (generic architecture \cite{zeng2025fair})}] (dm) {};
\end{scope}
\node[box, minimum width=20mm, fill=green!8] (out) at (17.9,0) {$\sigma(s_c(z_i))$\\[-1pt]\scriptsize hazard classes\\[-2pt]\scriptsize (multi-label)};
\draw[flow] (clf) -- (out);

\node[prop, minimum width=27mm] (tea) at (4.9,-3.6) {EMA teacher\\[-1pt]\scriptsize $\theta_{\text{teacher}}\leftarrow\alpha\theta_{\text{teacher}}+(1-\alpha)\theta_{\text{student}}$};
\node[prop, minimum width=26mm] (bx) at (8.6,-3.6) {Raw buffer $\mathcal{B}_x$\\[-1pt]\scriptsize few raw samples / class};
\node[prop, minimum width=26mm] (bz) at (11.6,-3.6) {Latent buffer $\mathcal{B}_z$\\[-1pt]\scriptsize many latents / class};
\node[prop, minimum width=30mm] (als) at (15.7,-3.6) {ALS-CBRS sampling\\[-1pt]\scriptsize $p=\min\{1,r_c\}\,\tilde\sigma(\mathrm{ALS}_c)$};
\coordinate (clarbot) at ($(als.south)+(0,-0.45)$);
\begin{scope}[on background layer]
  \node[draw=orange!70!black, rounded corners=3pt, fill=orange!4, fit=(tea)(bx)(bz)(als)(clarbot), inner sep=6pt,
        label={[font=\sffamily\small\bfseries, text=orange!50!black]below:CLAR components (proposed): dual replay buffer $\mathcal{M}$, ALS-CBRS, teacher}] (clar) {};
\end{scope}

\draw[->, thick, orange!70!black] (tea.north) to[out=70, in=200] node[lbl, pos=0.55, left, xshift=-1pt] {$\mathcal{L}_{\mathrm{KD}}$} (er.south west);
\draw[->, thick, gray!70!black] (clf.south) -- ++(0,-0.55) -| node[lbl, pos=0.28, above] {$(\mathbf{A}_c,\mathbf{v}_c)$, $z_i$} ([xshift=4mm]als.north);
\draw[->, thick, orange!70!black] (als.west) -- node[lbl, above] {admit $z_i$} (bz.east);
\draw[->, thick, orange!70!black] (als.south) -- ++(0,-0.4) -| node[lbl, pos=0.25, above] {admit $X_{i,t}$} (bx.south);
\draw[replay] (bx.north) -- node[lbl, right, align=left] {raw replay\\(re-encoded)} (er.south);
\draw[replay] (bz.north) to[out=90, in=225] node[lbl, right, pos=0.45, align=left] {latent\\replay} (clf.south west);
\end{tikzpicture}%
}
\caption{Overview of CLAR. After each computation task, the AI system in the MII yields the supervised learning performance $X_{i,t}^{(P)}$ and the runtime metrics $X_{i,t}^{(R)}$. The diagnosis model encodes both modalities into a latent representation $z_i$ and scores each hazard class with class-specific attention parameters $(\mathbf{A}_c,\mathbf{v}_c)$. The CLAR components (shaded) are the dual replay buffer, the ALS-guided class-balanced sampling, which uses $(\mathbf{A}_c,\mathbf{v}_c)$ to admit samples, and the EMA teacher, which anchors the latent space through $\mathcal{L}_{\mathrm{KD}}$. Dashed arrows denote replay.}
\label{fig:clar}
\end{figure*}
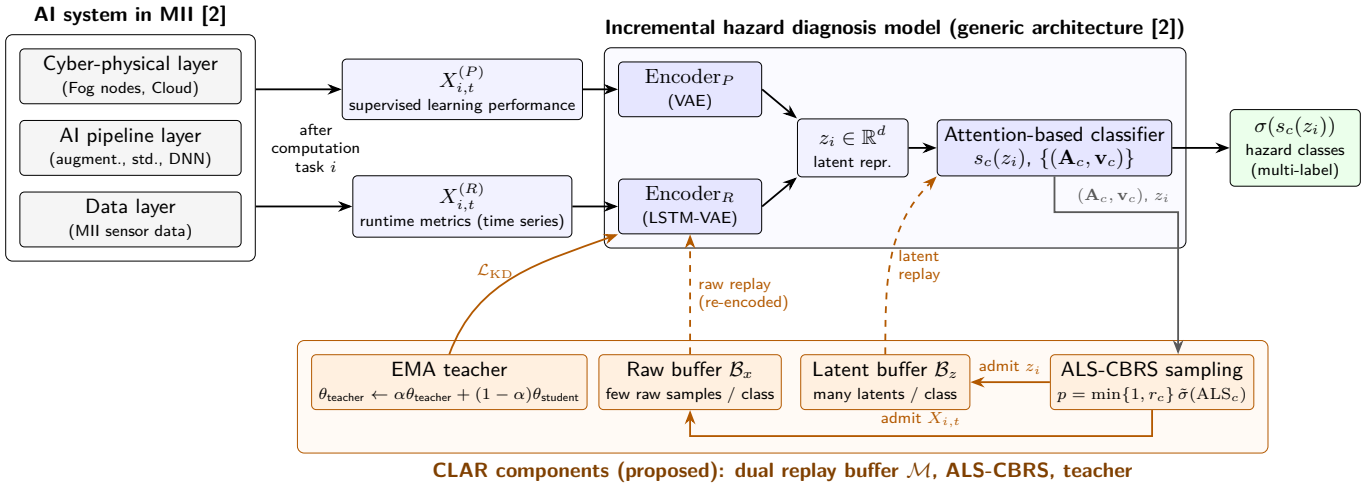

CLAR is a replay-based class-incremental framework.
Its key idea is to keep the memory of past hazards predominantly in the compact latent space of the diagnosis model, and to let the class-specific geometry learned by the classifier decide which samples are worth remembering.
Three components realize this idea, each addressing the assumptions in Section~\ref{sec:setting}, as shown in Fig.~\ref{fig:clar}: (i)
a \textit{latent-dominant dual replay buffer} (Section~\ref{sec:memory}) stores many compact latent representations and a few raw samples, which addresses the memory budget for high-dimensional samples (assumptions (ii) and (v));
(ii) an \textit{ALS-guided class-balanced sampling} strategy (Section~\ref{sec:als}) keeps the buffer balanced across hazard classes under an imbalanced, multi-label stream with recurring classes, and favors samples with a high ALS, i.e., samples that strongly drive the classifier response (assumptions (i), (iii), and (iv)); Section~\ref{sec:theory} analyzes when this selection improves replay;
(iii) a \textit{teacher-anchored latent space} (Section~\ref{sec:training}) constrains the encoders through distillation from a slowly updated teacher, which encourages stored latent representations to stay compatible with the evolving model.

The three components depend on each other.
The latent buffer stays useful only while the latent space remains close to the space in which its representations were stored, which is the role of the teacher, and the sampling strategy reads the parameters of the classifier, so the quality of memory selection improves as the classifier learns.
The diagnosis model itself follows the generic architecture of modality-specific encoders and class-specific attention used for root cause diagnosis in \cite{zeng2025fair}. CLAR applies to any encoder--classifier model with class-wise parameters.

\subsection{Incremental Hazard Diagnosis Model}\label{sec:model}

The incremental hazard diagnosis model has two stages: a multimodal feature extractor, which maps a raw sample to a low-dimensional latent representation, and an attention-based classifier, which maps the latent representation to hazard class scores.
This design follows the common strategy in multimodal learning of encoding each modality with a modality-specific encoder and fusing the results in a shared latent space \cite{ngiam2011multimodal, baltrusaitis2019multimodal, zeng2025fair}.

\textit{Multimodal feature extractor.}
Encoding into a low-dimensional space serves two purposes.
First, a raw sample is high-dimensional, and a classifier trained directly on raw samples from a few hazardous samples per class overfits easily.
Second, compact representations are cheap to store, which enables the latent buffer of Section~\ref{sec:memory}.
Modality-specific encoders map each modality to a latent representation,
\begin{equation}
z_i^{R}=\operatorname{Encoder}_{R}(X_{i,t}^{(R)}),\quad z_i^{R}\in\mathbb{R}^{d_R},
\end{equation}
\begin{equation}
z_i^{P}=\operatorname{Encoder}_{P}(X_{i,t}^{(P)}),\quad z_i^{P}\in\mathbb{R}^{d_P},
\end{equation}
which are concatenated into the latent representation of dimension $d=d_R+d_P$,
\begin{equation}
z_i=\operatorname{concat}(z_i^{R},z_i^{P})\in\mathbb{R}^{d}.
\label{eq:latent}
\end{equation}
Any modality-specific encoder can be used.
Following \cite{zeng2025fair}, we use variational autoencoders (VAEs) \cite{kingma2013auto}: an LSTM-based VAE for the time-series modality, whose step-wise outputs are aggregated over time, and a standard VAE for the tabular modality.
For modality $m\in\{R,P\}$, the encoder outputs a Gaussian approximate posterior $q_{\phi^{(m)}}(z_i^{m}\mid X_{i,t}^{(m)})$ with parameters $\phi^{(m)}$, whose mean is used as $z_i^{m}$; the decoders are used only for training (Section~\ref{sec:training}).

\textit{Attention-based classifier.}
For each seen class $c\in\mathcal{C}_t$, the classifier holds a trainable attention vector $\mathbf{v}_c\in\mathbb{R}^{d}$, which weights the latent dimensions relevant to class $c$, and a trainable attention matrix $\mathbf{A}_c\in\mathbb{R}^{d\times d}$, which models the interactions between them.
The logit of class $c$ is
\begin{equation}
s_c(z_i)=(\mathbf{A}_c z_i)^{\top}\big(\operatorname{diag}(\mathbf{v}_c)\,z_i\big),
\label{eq:clar_logit}
\end{equation}
where $\operatorname{diag}(\mathbf{v}_c)\in\mathbb{R}^{d\times d}$ is the diagonal matrix formed by $\mathbf{v}_c$, and the predicted probability of class $c$ is $\sigma(s_c(z_i))$ with the sigmoid function $\sigma(\cdot)$.
Each class is scored with its own parameters, which suits the multi-label setting and mirrors the root-cause-specific attention heads of \cite{zeng2025fair}; a new class is accommodated by adding a pair $(\mathbf{A}_c,\mathbf{v}_c)$ when it first appears.
We refer to $\{(\mathbf{A}_c,\mathbf{v}_c)\}_{c\in\mathcal{C}_t}$ as the \textit{learned attention geometry}, since it determines how the classifier responds to each region of the latent space; CLAR reuses it to select samples for the buffer (Section~\ref{sec:als}).

\subsection{Latent-Dominant Dual Replay Buffer}\label{sec:memory}

In replay-based CIL, a memory buffer is a fixed-capacity store of past training samples that are trained together with the data of each new task (Section~\ref{sec:related}).
Under a strict memory budget, storing raw samples alone leaves very few exemplars per hazard class: in our case study, one raw sample is about 260 times larger than the 150 latent representations of the latent buffer combined.
Storing latent representations alone is insufficient as well, because the encoders keep being updated and need raw inputs of earlier classes to preserve their representation.

CLAR therefore maintains a dual replay buffer $\mathcal{M}=\{\mathcal{B}_x,\mathcal{B}_z\}$ with two complementary roles.
The latent buffer $\mathcal{B}_z$ stores many latent representations $(z_i,y_i)$ and gives the classifier broad coverage of each earlier class at low cost.
The raw buffer $\mathcal{B}_x$ stores a few raw samples $(X_{i,t},y_i)$, which are re-encoded by the current encoders during training and keep the encoders exposed to earlier classes.
To keep the buffer class-balanced, both $\mathcal{B}_x$ and $\mathcal{B}_z$ are partitioned by hazard class into $\mathcal{B}_x[c]$ and $\mathcal{B}_z[c]$, with per-class quotas $q_{c,\text{raw}}=\lfloor|\mathcal{B}_x|_{\max}/|\mathcal{C}_t|\rfloor$ and $q_{c,\text{lat}}=\lfloor|\mathcal{B}_z|_{\max}/|\mathcal{C}_t|\rfloor$, where $|\cdot|_{\max}$ denotes the capacity.
The quotas are recomputed whenever new classes appear; the floor can leave up to $|\mathcal{C}_t|-1$ slots of each buffer unused, and positive quotas require $|\mathcal{C}_t|$ to stay below the capacity.
We call the buffer latent-dominant because latent representations make up most of its slots; in bytes, the few raw samples still dominate.
Only hazardous samples are stored, since normal samples are abundant in every task, and a multi-label sample can belong to several partitions at the same time.

\subsection{ALS-Guided Class-Balanced Sampling}\label{sec:als}

The sampling strategy, referred to as ALS-CBRS, decides which samples enter each partition of the buffer.
While a partition is below its quota, every new sample of that class is inserted.
Once the partition is full, a new sample of class $c$ replaces a randomly chosen stored sample with probability
\begin{equation}
p=\min\{1,r_c\}\cdot\tilde{\sigma}\big(\operatorname{ALS}_c(z_i)\big), \qquad r_c=\frac{q_c}{N_c},
\label{eq:pinsert}
\end{equation}
where $q_c$ is the quota of the corresponding buffer and $N_c$ is the cumulative number of class-$c$ samples observed so far.

The reservoir factor $r_c$ is the admission probability of reservoir sampling \cite{vitter1985random} applied within each class, as in CBRS \cite{chrysakis2020online}.
With $\tilde{\sigma}\equiv1$, it keeps each partition a uniform random sample of the class-$c$ samples observed so far.
The second factor scales this probability by the response of the classifier to the sample, measured by the Attention Leverage Score
\begin{equation}
\operatorname{ALS}_c(z_i)=\sqrt{\big(\operatorname{diag}(\mathbf{v}_c)z_i\big)^{\top}P_c\big(\operatorname{diag}(\mathbf{v}_c)z_i\big)},
\label{eq:als}
\end{equation}
where $P_c=\mathbf{A}_c\mathbf{A}_c^{\top}+\varepsilon\mathbf{I}$, and $\varepsilon>0$ makes $P_c$ positive definite.
The term $\operatorname{diag}(\mathbf{v}_c)z_i$ keeps the latent dimensions that class $c$ attends to, and $P_c$ weights them by the interactions learned in $\mathbf{A}_c$, so the ALS is a Mahalanobis-type norm in the metric induced by the attention geometry of class $c$.

The ALS is directly tied to the classifier.
Rewriting Eq.~\eqref{eq:clar_logit} as $s_c(z_i)=z_i^{\top}\mathbf{A}_c^{\top}\operatorname{diag}(\mathbf{v}_c)z_i$ and applying the Cauchy--Schwarz inequality gives
\begin{equation}
|s_c(z_i)|\le\|z_i\|_2\,\big\|\mathbf{A}_c^{\top}\operatorname{diag}(\mathbf{v}_c)z_i\big\|_2\le\|z_i\|_2\,\operatorname{ALS}_c(z_i),
\label{eq:als_logit_bound}
\end{equation}
where the second inequality holds because $\operatorname{ALS}_c(z_i)^2=\|\mathbf{A}_c^{\top}\operatorname{diag}(\mathbf{v}_c)z_i\|_2^2+\varepsilon\|\operatorname{diag}(\mathbf{v}_c)z_i\|_2^2$.
The bound holds for every sample, every class, and every $\varepsilon>0$, with near-equality when $z_i$ is aligned with $\mathbf{A}_c^{\top}\operatorname{diag}(\mathbf{v}_c)z_i$ and $\varepsilon$ is small.
For a given sample norm, a large logit magnitude for class $c$ thus requires a large ALS, so the ALS measures how strongly a sample can drive the class-$c$ output in the classifier's own geometry.

Finally, the scaled and shifted sigmoid $\tilde{\sigma}(u)=2/(1+e^{-u})-1$ maps the non-negative ALS to $[0,1)$.
The standard sigmoid maps non-negative inputs to $[0.5,1)$, whereas $\tilde{\sigma}$ lets weakly responding samples have an admission probability close to zero.
The two factors play distinct roles.
The per-class quotas keep the buffer balanced across hazard classes for any admission factor (Proposition~\ref{prop:balance}), and the ALS factor tilts the content of each partition toward samples with a high ALS.
Section~\ref{sec:theory} characterizes the resulting buffer and derives when this tilt improves replay over class-balanced uniform replay.
It is applied to $\mathcal{B}_x$ and $\mathcal{B}_z$ with their respective quotas at the end of each task, using the updated encoders and classifier (Algorithm~\ref{alg:alscbrs}). 

\begin{algorithm}[t]
\caption{ALS-CBRS buffer update after task $t$}
\label{alg:alscbrs}
\begin{algorithmic}[1]
\Require $\mathcal{M}=\{\mathcal{B}_x,\mathcal{B}_z\}$, $\mathcal{D}_t^{\text{train}}$, $\mathcal{C}_{t-1}$, $\mathcal{C}_{\mathrm{new}}$, $\{(\mathbf{A}_c,\mathbf{v}_c)\}$
\If{$\mathcal{C}_{\mathrm{new}}\neq\emptyset$}
  \State $\mathcal{C}_t\gets\mathcal{C}_{t-1}\cup\mathcal{C}_{\mathrm{new}}$; recompute $q_{c,\text{raw}},q_{c,\text{lat}}$
  \State Evict random items from partitions over their quota
\EndIf
\ForAll{hazardous $(X_{i,t},y_i)\in\mathcal{D}_t^{\text{train}}\setminus\mathcal{D}_t^{\text{retrieve}}$}
  \State $z_i\gets\operatorname{concat}(\operatorname{Encoder}_R(X_{i,t}^{(R)}),\operatorname{Encoder}_P(X_{i,t}^{(P)}))$
  \ForAll{$c\in\mathcal{C}_t$ with $y_{i,c}=1$}
    \State $N_c\gets N_c+1$
    \ForAll{$\mathcal{B}\in\{\mathcal{B}_x,\mathcal{B}_z\}$}
      \State $(e,q_c)\gets(X_{i,t},q_{c,\text{raw}})$ if $\mathcal{B}=\mathcal{B}_x$, else $(z_i,q_{c,\text{lat}})$
      \State Draw $u\sim\mathrm{Uniform}(0,1)$
      \If{$|\mathcal{B}[c]|<q_c$}
        \State Insert $(e,y_i)$ into $\mathcal{B}[c]$
      \ElsIf{$u<\min\{1,q_c/N_c\}\cdot\tilde{\sigma}(\operatorname{ALS}_c(z_i))$}
        \State Replace a random item of $\mathcal{B}[c]$ with $(e,y_i)$
      \EndIf
    \EndFor
  \EndFor
\EndFor
\end{algorithmic}
\end{algorithm}

\subsection{Teacher-Anchored Training Objective}\label{sec:training}

When the encoders are updated on new hazard classes, the latent space shifts.
This representation drift has two consequences: the classifier loses the representation of earlier classes, and the latent representations in $\mathcal{B}_z$, which were produced by earlier encoders, drift away from the representations that the current encoders produce.
CLAR counters the drift with a teacher model whose parameters $\theta_{\text{teacher}}$ are an exponential moving average (EMA) of the parameters $\theta_{\text{student}}$ of the diagnosis model (the student) \cite{tarvainen2017mean}.
After training on task $t$, the teacher is updated as
\begin{equation}
\theta_{\text{teacher}}^{(t)}=\alpha\,\theta_{\text{teacher}}^{(t-1)}+(1-\alpha)\,\theta_{\text{student}}^{(t)},
\label{eq:ema}
\end{equation}
with momentum coefficient $\alpha\in[0,1]$, so the teacher changes slowly and provides a stable reference for the latent space.
The student is trained on the training pool $\mathcal{D}_t^{\text{train}}\cup\mathcal{D}_t^{\text{retrieve}}$, where $\mathcal{D}_t^{\text{retrieve}}$ is a class-balanced set of raw samples retrieved from $\mathcal{B}_x$, with three loss terms.

\textit{Weighted multi-label classification loss.}
The classifier is trained on the classification pool $\mathcal{S}_{\mathrm{cls}}$, which contains the latent representations $z_i^{\text{train}}$ of the current mini-batch, encoded by the current encoders, and an equal number of latent representations $z_j^{\text{retrieved}}$ retrieved from $\mathcal{B}_z$.
To counter class imbalance, a weighted binary cross-entropy is used,
\begin{equation}
\mathcal{L}_{\mathrm{clf}}=-\frac{1}{|\mathcal{S}_{\mathrm{cls}}|}\sum_{(z_i,y_i)\in\mathcal{S}_{\mathrm{cls}}}\ \sum_{c\in\mathcal{C}_t}\ell_{i,c},
\end{equation}
\begin{equation}
\ell_{i,c}=w_c\,y_{i,c}\log\sigma(\hat{y}_{i,c})+(1-y_{i,c})\log\big(1-\sigma(\hat{y}_{i,c})\big),
\label{eq:ell}
\end{equation}
where $\hat{y}_{i,c}=s_c(z_i)$, and the positive weight $w_c$ is the ratio of negative to positive samples of class $c$ in the training pool of task $t$.

\textit{VAE loss.}
The encoders and decoders are trained with the negative evidence lower bound of each modality, which consists of a reconstruction loss and a Kullback--Leibler (KL) divergence toward the standard normal prior \cite{kingma2013auto, zeng2025fair},
\begin{equation}
\begin{aligned}
\mathcal{L}_{\mathrm{VAE}}=\sum_{m\in\{R,P\}}\Big(&\mathcal{L}_{\mathrm{recon}}\big(X_{i,t}^{(m)},\hat{X}_{i,t}^{(m)}\big)\\
&+\mathrm{KL}\big(q_{\phi^{(m)}}(z_i^{m}\mid X_{i,t}^{(m)})\,\|\,\mathcal{N}(\mathbf{0},\mathbf{I})\big)\Big),
\end{aligned}
\end{equation}
averaged over the training pool, where $\hat{X}_{i,t}^{(m)}$ is the reconstruction by the decoder of modality $m$.
This term keeps the latent space informative about both modalities and, through the retrieved raw samples, about earlier hazard classes.

\textit{Latent distillation loss.}
Knowledge distillation \cite{hinton2015distilling, li2017learning} is applied in the latent space of each encoder by penalizing the divergence between the posteriors of the student and the teacher,

\begin{equation}
\begin{aligned}
\mathcal{L}_{\mathrm{KD}} = \sum_{m\in\{R,P\}} \mathrm{KL}\Big(
&q_{\phi_{\text{student}}^{(m)}}(z_i^{m}\mid X_{i,t}^{(m)}) \\
&\parallel q_{\phi_{\text{teacher}}^{(m)}}(z_i^{m}\mid X_{i,t}^{(m)})
\Big),
\end{aligned}
\end{equation}
averaged over the training pool.

\textit{Overall objective.}
The student minimizes
\begin{equation}
\mathcal{L}_{\mathrm{total}}=\mathcal{L}_{\mathrm{clf}}+\lambda_{\mathrm{VAE}}\,\mathcal{L}_{\mathrm{VAE}}+\lambda_{\mathrm{KD}}\,\mathcal{L}_{\mathrm{KD}},
\label{eq:total}
\end{equation}
where $\lambda_{\mathrm{VAE}}\ge0$ and $\lambda_{\mathrm{KD}}\ge0$ balance representation learning and stability against diagnosis accuracy.

\subsection{Incremental Training Procedure}\label{sec:procedure}

On the base task $\mathcal{D}_0$, the diagnosis model is trained with $\mathcal{L}_{\mathrm{clf}}+\lambda_{\mathrm{VAE}}\mathcal{L}_{\mathrm{VAE}}$, the teacher is initialized as a copy of the trained model, and the buffer is filled with Algorithm~\ref{alg:alscbrs}.
Each incremental task then follows Algorithm~\ref{alg:train}.
Raw samples are retrieved once per task and appended to the training set, so they contribute to all three loss terms; latent representations are retrieved for every mini-batch and contribute to $\mathcal{L}_{\mathrm{clf}}$.

\begin{algorithm}[t]
\caption{Incremental training of CLAR on task $t>0$}
\label{alg:train}
\begin{algorithmic}[1]
\State Retrieve class-balanced $\mathcal{D}_t^{\text{retrieve}}$ from $\mathcal{B}_x$; $\ \mathcal{D}_t^{\text{train}}\gets\mathcal{D}_t^{\text{train}}\cup\mathcal{D}_t^{\text{retrieve}}$
\State Add $(\mathbf{A}_c,\mathbf{v}_c)$ for each $c\in\mathcal{C}_{\mathrm{new}}$
\Repeat\ for each mini-batch of size $b$ from $\mathcal{D}_t^{\text{train}}$
  \State Encode the mini-batch into $z_i^{\text{train}}$ by Eq.~\eqref{eq:latent}
  \State $\mathcal{S}_{\mathrm{cls}}\gets\{z_i^{\text{train}}\}\cup$ class-balanced $\{z_j^{\text{retrieved}}\}_{j=1}^{b}$ from $\mathcal{B}_z$
  \State Update $\theta_{\text{student}}$ by a gradient step on $\mathcal{L}_{\mathrm{total}}$ in Eq.~\eqref{eq:total}
\Until{early stopping}
\State Update the teacher by Eq.~\eqref{eq:ema}; update $\mathcal{M}$ by Algorithm~\ref{alg:alscbrs}
\end{algorithmic}
\end{algorithm}

\section{Theoretical Analysis of ALS-CBRS}\label{sec:theory}

This section analyzes, for an idealized model of the buffer update and of training, what ALS-CBRS stores and when its selection benefits learning.
The results describe a mechanism that can explain the role of ALS-guided replay, which provide guarantees for this model. The full adaptive training procedure of CLAR is assessed empirically in Section~\ref{sec:exp}.
We consider one partition $\mathcal{B}[c]$, which can be $\mathcal{B}_x[c]$ or $\mathcal{B}_z[c]$, with quota $q_c$, and compare ALS-CBRS with its uniform counterpart, i.e., class-balanced reservoir sampling with the same quota, obtained by setting $\tilde{\sigma}\equiv1$ in Eq.~\eqref{eq:pinsert} and referred to as CBRS below \cite{chrysakis2020online}.
Proofs are given in the Appendix.

\subsection{Class Balance}
\begin{proposition}[Class balance]\label{prop:balance}
Under Algorithm~\ref{alg:alscbrs}, after every buffer update, $|\mathcal{B}[c]|=\min\{q_c,N_c\}$ for every $c\in\mathcal{C}_t$ and for both $\mathcal{B}_x$ and $\mathcal{B}_z$.
\end{proposition}
Every class observed at least $q_c$ times thus occupies exactly $q_c$ slots, however frequent it is in the stream, and a rarer class keeps all of its $N_c$ samples.
Class balance therefore follows from the per-class quotas and holds for any admission factor, including $\tilde{\sigma}(\operatorname{ALS}_c)$. 
The ALS factor determines which samples occupy the slots.
For multi-label data, balance refers to the partition sizes: a sample in $\mathcal{B}[c]$ can carry further labels, so the label counts over the whole buffer still reflect co-occurrence.

\subsection{Replay Law of ALS-CBRS}
Fix class $c$, a quota $q_c\ge1$, and the parameters used by Algorithm~\ref{alg:alscbrs}, i.e., the encoders and $(\mathbf{A}_c,\mathbf{v}_c)$.
We treat these parameters, and hence the gate below, as fixed and independent of the arrivals analyzed; in Algorithm~\ref{alg:alscbrs} the gate is learned partly from the same task's samples that it scores, so this independence is an idealization (Section~\ref{sec:theory}-D).
Let the class-$c$ samples arrive as independent and identically distributed (i.i.d.) pairs $(z,y)\sim\pi_c$, where $z$ is the latent representation and $y$ the label vector, and let $N_c$ count the arrivals.
Define the ALS gate and its mean as
\begin{equation}
h_c(z)=\tilde{\sigma}\big(\operatorname{ALS}_c(z)\big)\in[0,1),\qquad \bar{h}_c=\mathbb{E}_{\pi_c}h_c(z)>0.
\label{eq:gate}
\end{equation}
By Eq.~\eqref{eq:pinsert}, the first $q_c$ arrivals fill the partition, and afterward the $N_c$-th arrival replaces a uniformly chosen entry with probability $(q_c/N_c)\,h_c(z)$.
The ALS-weighted class distribution $\tilde{\pi}_c$ is defined by
\begin{equation}
d\tilde{\pi}_c=\frac{h_c}{\bar{h}_c}\,d\pi_c .
\label{eq:tilt}
\end{equation}

\begin{lemma}[Replay law]\label{lem:law}
Every sample admitted after the partition is full has law $\tilde{\pi}_c$.
Each entry of the initial fill survives the first $N_c$ arrivals with probability
\begin{equation}
\kappa_c(N_c)=\prod_{j=q_c+1}^{N_c}\Big(1-\frac{\bar{h}_c}{j}\Big),
\label{eq:retention}
\end{equation}
and the total variation distance between the law of $\mathcal{B}[c]$ and $\tilde{\pi}_c^{\otimes q_c}$ is at most $\min\{1,q_c\kappa_c(N_c)\}$.
In particular, $\mathcal{B}[c]$ converges in distribution to $q_c$ i.i.d.\ draws from $\tilde{\pi}_c$ as $N_c\to\infty$.
\end{lemma}

Lemma~\ref{lem:law} states precisely what favoring high-ALS samples means: in the long run, each partition is a sample from the class distribution reweighted by $\tilde{\sigma}(\operatorname{ALS}_c)$, whereas CBRS ($h_c\equiv1$) keeps a uniform sample of $\pi_c$.
For finite streams, entries of the initial fill, which enter without ALS, persist with probability $\kappa_c(N_c)$.
CBRS gives $\kappa_c(N_c)=q_c/N_c$, and ALS-CBRS gives $\kappa_c(N_c)\approx(q_c/N_c)^{\bar{h}_c}$ for large $q_c$, so a gate with a small mean $\bar{h}_c$ refreshes the partition more slowly.
With only a few samples per hazard class, $N_c/q_c$ is small and part of each partition originates from the initial fill; the results below hold at every finite $N_c$ and account for this effect.

\subsection{Replay-Gradient Error and Margin Stability}
To relate the stored samples to learning, consider an output class $c'\in\mathcal{C}_t$, for instance $c'=c$, and hold $\mathbf{A}_{c'}$ and the encoders fixed.
The logit in Eq.~\eqref{eq:clar_logit} is then linear in the attention vector,
\begin{equation}
s_{c'}(z)=\mathbf{v}^{\top}\psi_{c'}(z),\qquad \psi_{c'}(z)=z\odot(\mathbf{A}_{c'}z),
\label{eq:psi}
\end{equation}
where $\mathbf{v}$ denotes the attention vector $\mathbf{v}_{c'}$ treated as the optimization variable.
Let $-\ell_{c'}(\mathbf{v};z,y)$ be the weighted binary cross-entropy of Eq.~\eqref{eq:ell} for class $c'$ with logit $\mathbf{v}^{\top}\psi_{c'}(z)$ and a fixed weight $w_{c'}$.
The population and replay objectives are
\begin{equation}
\begin{aligned}
\mathcal{J}(\mathbf{v})&=\mathbb{E}_{\pi_c}\big[-\ell_{c'}(\mathbf{v};z,y)\big]+\Omega(\mathbf{v}),\\
\hat{\mathcal{J}}(\mathbf{v})&=\frac{1}{q_c}\sum_{(z,y)\in\mathcal{B}[c]}\big[-\ell_{c'}(\mathbf{v};z,y)\big]+\Omega(\mathbf{v}),
\end{aligned}
\label{eq:objectives}
\end{equation}
where $\Omega$ is a twice differentiable convex term that collects the remaining training loss, such as the loss on the current task, held fixed and common to both sampling rules. \yz{Is any weight decay or L2 regularization used in training? If yes, cite it here as a source of the curvature assumed in Theorem~\ref{thm:main}.}
The replay term contributes the Hessian $q_c^{-1}\sum_{(z,y)\in\mathcal{B}[c]}a\,\sigma(1-\sigma)\,\psi_{c'}(z)\psi_{c'}(z)^{\top}$ with $a=w_{c'}y_{c'}+1-y_{c'}$ and $\sigma=\sigma(\mathbf{v}^{\top}\psi_{c'}(z))$, whose rank is at most $q_c$; for $q_c<d$, the curvature condition of Theorem~\ref{thm:main} therefore relies on $\Omega$, and we keep it as an assumption of the analysis.
Let $\mathbf{v}^{\star}$ and $\hat{\mathbf{v}}$ minimize $\mathcal{J}$ and $\hat{\mathcal{J}}$, respectively.
The replay gradient at $\mathbf{v}^{\star}$ is governed by
\begin{equation}
\mathbf{g}(z,y)=\big(w_{c'}y_{c'}+1-y_{c'}\big)\big(\sigma(\mathbf{v}^{\star\top}\psi_{c'}(z))-y_{c'}\big)\,\psi_{c'}(z),
\label{eq:grad}
\end{equation}
with the statistics
\begin{equation}
\begin{gathered}
V_c=\operatorname{tr}\operatorname{Cov}_{\pi_c}(\mathbf{g}),\qquad \tilde{V}_c=\operatorname{tr}\operatorname{Cov}_{\tilde{\pi}_c}(\mathbf{g}),\\
\boldsymbol{\delta}_c=\mathbb{E}_{\tilde{\pi}_c}\mathbf{g}-\mathbb{E}_{\pi_c}\mathbf{g}=\operatorname{Cov}_{\pi_c}(\mathbf{g},h_c)/\bar{h}_c .
\end{gathered}
\label{eq:stats}
\end{equation}
Here, $V_c$ and $\tilde{V}_c$ are the gradient variances under uniform and ALS-weighted selection, and $\boldsymbol{\delta}_c$ is the selection bias.

\begin{theorem}[Replay-gradient error and margin stability]\label{thm:main}
Suppose $\mathbb{E}_{\pi_c}\|z\|_2^4<\infty$, $\mathbf{v}^{\star}$ and $\hat{\mathbf{v}}$ are interior minimizers, and $\nabla^2\hat{\mathcal{J}}\succeq\mu\mathbf{I}$ on the segment between $\mathbf{v}^{\star}$ and $\hat{\mathbf{v}}$ almost surely, with a deterministic $\mu>0$ common to both sampling rules.
Let $\bar{\mathbf{g}}=q_c^{-1}\sum_{(z,y)\in\mathcal{B}[c]}\mathbf{g}(z,y)$, and let $\mathcal{E}_c(N_c)=\mathbb{E}\|\bar{\mathbf{g}}-\mathbb{E}_{\pi_c}\mathbf{g}\|_2^2$ be the mean-squared error of the replay gradient of ALS-CBRS after $N_c$ arrivals and $\mathcal{E}_c^{\mathrm{CBRS}}(N_c)$ that of CBRS. Then
\begin{equation}
\lim_{N_c\to\infty}\mathcal{E}_c(N_c)=\frac{\tilde{V}_c}{q_c}+\|\boldsymbol{\delta}_c\|_2^2,\qquad
\mathcal{E}_c^{\mathrm{CBRS}}(N_c)=\frac{V_c}{q_c}\ \ \forall N_c\ge q_c .
\label{eq:limit}
\end{equation}
Moreover, for a test sample $(z_*,y_*)$ and a decision threshold $\tau$, the margin $\gamma(\mathbf{v})=(2y_{*,c'}-1)\big(\mathbf{v}^{\top}\psi_{c'}(z_*)-\tau\big)$ satisfies
\begin{equation}
\mathbb{E}\big(\gamma(\hat{\mathbf{v}})-\gamma(\mathbf{v}^{\star})\big)^2\le\frac{\|\psi_{c'}(z_*)\|_2^2}{\mu^2}\,\mathcal{E}_c(N_c),
\label{eq:margin}
\end{equation}
and the same bound with $\mathcal{E}_c^{\mathrm{CBRS}}(N_c)$ holds for CBRS.
\end{theorem}

\begin{corollary}[Critical capacity]\label{cor:qstar}
Under Theorem~\ref{thm:main}, suppose $\boldsymbol{\delta}_c\neq\mathbf{0}$ and define
\begin{equation}
q_c^{\star}=\frac{V_c-\tilde{V}_c}{\|\boldsymbol{\delta}_c\|_2^2}
=1-\frac{\operatorname{Cov}_{\pi_c}\big(h_c,\|\mathbf{g}-\mathbb{E}_{\pi_c}\mathbf{g}\|_2^2\big)}{\bar{h}_c\|\boldsymbol{\delta}_c\|_2^2}.
\label{eq:qstar}
\end{equation}
If $1\le q_c<q_c^{\star}$, then $\mathcal{E}_c(N_c)<\mathcal{E}_c^{\mathrm{CBRS}}(N_c)$ for every $N_c>q_c$, and the margin bound in Eq.~\eqref{eq:margin} is smaller for ALS-CBRS whenever $\psi_{c'}(z_*)\neq\mathbf{0}$.
If $q_c>q_c^{\star}$, then $\mathcal{E}_c(N_c)>\mathcal{E}_c^{\mathrm{CBRS}}(N_c)$ for all sufficiently large $N_c$.
If $\boldsymbol{\delta}_c=\mathbf{0}$, then $\mathcal{E}_c(N_c)<\mathcal{E}_c^{\mathrm{CBRS}}(N_c)$ for every $N_c>q_c$ exactly when $\tilde{V}_c<V_c$.
\end{corollary}

\subsection{Implications for CLAR}
Theorem~\ref{thm:main} splits the replay-gradient error into a sampling variance $\tilde{V}_c/q_c$ and a squared selection bias $\|\boldsymbol{\delta}_c\|_2^2$.
CBRS is unbiased, and ALS-CBRS trades a bias for a change in variance.
By Corollary~\ref{cor:qstar}, in the limit $N_c\to\infty$, a capacity $q_c\ge1$ with an advantage over CBRS exists exactly when $q_c^{\star}>1$ (the exact finite-stream criterion is given in Appendix~\ref{app:cor}), i.e., when $\operatorname{Cov}_{\pi_c}(h_c,\|\mathbf{g}-\mathbb{E}_{\pi_c}\mathbf{g}\|_2^2)<0$: the ALS gate has to favor samples whose replay gradients lie close to the class-average gradient.
This is the precise sense in which ALS-selected samples are informative for replay: they represent the gradient of their class with less variance.
Lemma~\ref{lem:law}, Theorem~\ref{thm:main}, and Corollary~\ref{cor:qstar} hold for any fixed gate with values in $[0,1]$ and a positive mean.
The logit bound in Eq.~\eqref{eq:als_logit_bound} motivates the ALS gate and leaves the sign of the covariance open, so the analysis characterizes when gated replay helps, and whether the ALS gate meets the condition is a property of the data and the learned attention geometry that can be estimated at training checkpoints.
Through Eq.~\eqref{eq:margin}, a smaller replay-gradient error gives a smaller upper bound on the expected squared deviation of the classification margin of each test sample, which relates the sampling rule to the stability of the decision boundaries of earlier hazard classes.

\textit{Per-class capacity and the dual buffer.}
Holding the gradient statistics fixed, the limiting advantage over CBRS, $(V_c-\tilde{V}_c)/q_c-\|\boldsymbol{\delta}_c\|_2^2$, grows as $q_c$ decreases.
In CLAR, the quotas $q_{c,\text{raw}}=\lfloor|\mathcal{B}_x|_{\max}/|\mathcal{C}_t|\rfloor$ and $q_{c,\text{lat}}=\lfloor|\mathcal{B}_z|_{\max}/|\mathcal{C}_t|\rfloor$ shrink as hazard classes accumulate, so ALS selection matters most when many classes share a small buffer.
The two buffers play complementary roles: the latent-dominant design enlarges $q_{c,\text{lat}}$ and thereby reduces the variance term under any sampling rule, while ALS selection gains most in partitions with a small quota, such as those of the raw buffer.

\textit{Scope.}
The analysis holds the encoders, $\mathbf{A}_{c'}$, and the ALS gate fixed and independent of the analyzed arrivals, assumes i.i.d.\ arrivals within a class, a fixed quota $q_c\ge1$ and class weight, and a curvature bound common to both rules, optimizes one attention vector with the remaining loss held fixed, and compares ALS-CBRS with CBRS at the same quota.
CLAR departs from this model in four ways: the gate is learned partly from the samples it scores; the parameters change between tasks, a drift that the teacher-anchored latent space (Section~\ref{sec:training}) is designed to slow; the whole network is trained jointly with several interacting partitions, and multi-label samples couple the partitions of co-occurring classes; and quotas change when classes appear, where random eviction keeps a uniformly chosen subset and thus preserves the composition of a partition in expectation.
The analysis concerns the replay gradient and the margin deviation, and the diagnosis performance along the incremental sequence is assessed empirically in Section~\ref{sec:exp}.

\section{Experiments}\label{sec:exp}

\subsection{MII Testbed and AI System}
We evaluate CLAR on hazard diagnosis data from the MII testbed of \cite{zeng2025fair}, whose AI system provides product quality modeling for an Aerosol Jet Printing (AJP) process as its computation service.
The manufacturing data come from a real AJP dataset of 95 samples, each with six in situ process variables in time-series format (atomizer gas flow, sheath gas flow, current, nozzle X-coordinate, nozzle Y-coordinate, and nozzle vibration) and a binary quality label (conforming or nonconforming).
Four synthetic time-series variables with negligible predictive power are added, giving ten input variables.
Five AJP machines are simulated, each with its own ground-truth DNN model that labels samples generated by augmenting the 95 original samples (time warping, pooling, convolution, and others), which yields 500 samples per machine.
A multivariate time-series DNN pipeline \cite{shojaee2021deep} is trained for each machine, and the top three pipelines of each machine are deployed on five fog nodes (Raspberry Pi computers) coordinated by a Cloud server.
A computation task is the quality prediction for a batch of data from one machine, executed on one computation node.

Hazards are injected by varying the levels of the root cause factors of \cite{zeng2025fair} in the data and AI pipeline layers, as summarized in Table~\ref{tab:hazard_classes}.
Hazards in the cyber-physical layer, i.e., failed fog nodes and communication channels, prevent the computation task from completing and leave the supervised learning performance missing, so they are trivially identifiable \cite{zeng2025fair}; we therefore consider the five root cause factors whose computation tasks complete successfully.

\begin{table}[htbp]
\centering
\caption{Hazard classes and the corresponding root cause factors \cite{zeng2025fair}.}
\label{tab:hazard_classes}
\begin{tabularx}{\linewidth}{l|l|>{\raggedright\arraybackslash}X}
\hline
\textbf{Class} & \textbf{Layer} & \textbf{Root cause factor and simulation} \\ \hline
Class 0 & Data & Sensor contamination or failure: \% of contaminated or failed sensors among the ten inputs \\ \hline
Class 1 & Data & Low signal-to-noise ratio (SNR): random noise injected at controlled scales \\ \hline
Class 2 & Data & Distribution change of input variables: Gaussian mixture data with controlled KL divergence \\ \hline
Class 3 & Data & Class imbalance: nonconforming-to-conforming ratio within a data batch \\ \hline
Class 4 & AI pipeline & Pipeline singularity: \% of pipelines forced to produce identical predictions \\ \hline
\end{tabularx}
\end{table}

\subsection{Diagnosis Dataset}
The diagnosis dataset consists of 4,705 successfully executed computation tasks, each forming one sample $X_{i,t}=\{X_{i,t}^{(P)},X_{i,t}^{(R)}\}$ as defined in Section~\ref{sec:setting}.
The supervised learning performance $X_{i,t}^{(P)}$ is a 92-dimensional vector that concatenates 9 core performance metrics (accuracy, precision, and F1 score of the top three pipelines) with 83 additional features. 
The runtime metrics $X_{i,t}^{(R)}$ form a tensor of shape $500\times500\times10$, which records 10 hardware variables, including CPU temperature, memory consumption, and data transmission volume, over 500 time stamps for the 500 data samples of the processed batch.
Each sample is labeled with the hazard classes in Table~\ref{tab:hazard_classes}, and several hazards can co-occur in the same computation task.

\subsection{Class-Incremental Protocol}
The dataset is highly imbalanced and sparse: only 273 of the 4,705 samples contain at least one hazard, and the remaining samples represent normal operation.
The base task is built from the most frequent label configuration to initialize the feature extractor and the classifier.
Among all three-label subsets, $\{\text{Class } 0,\text{Class } 3,\text{Class } 4\}$ has the largest support, and the base task contains all hazardous samples of these classes. \yz{Explain what happens to samples labeled with a base class (0, 3, 4) together with Class 1 or 2: are they excluded, moved to the task that introduces Class 1 or 2, or kept in the base task with the future label masked? Otherwise Classes 1 and 2 may appear in the base task.}
The remaining classes enter the incremental tasks in the order given by the class schedule $\mathcal{S}_{\text{sched}}=[2, 1]$, i.e., Class 2 appears before Class 1. \yz{Report the train/test split of each task (ratio, stratified or not, any grouping by machine or scenario) and the preprocessing (e.g., normalization fitted on which data).}

The task-allocation hyperparameter $K=40$ sets the number of hazardous samples per incremental task, which yields five incremental tasks and six tasks in total.
Since only two new classes appear after the base task, most incremental tasks introduce zero new classes and revisit classes that have already been seen, which reflects the sparse and irregular emergence of new hazards in the MII (assumption (iv)).
Normal samples are added to each task according to the imbalance ratio, i.e., the number of normal samples per hazardous sample, which is $\rho_{\text{base}}=10$ for the base task and $\rho_{\text{inc}}\in\{10,15\}$ for the incremental tasks.

\subsection{Implementation Details}
CLAR follows Algorithm~\ref{alg:train} with mini-batch size $b=8$, which is small owing to the size of the raw samples.
The loss weights in Eq.~\eqref{eq:total} are $\lambda_{\mathrm{VAE}}=0.1$ and $\lambda_{\mathrm{KD}}=1.0$, and the EMA momentum in Eq.~\eqref{eq:ema} is $\alpha=0.6$.
Training on each task stops early when the relative change in $\mathcal{L}_{\mathrm{clf}}$ falls below $0.005$ for 5 epochs without improvement.

All replay methods are configured with comparable memory budgets.
The baselines have a capacity of 50 raw samples and retrieve 40 of them per task.
CLAR has a capacity of 49 raw samples in $\mathcal{B}_x$ and 150 latent representations of dimension $d=64$ in $\mathcal{B}_z$; with five classes, $q_{c,\text{raw}}=9$ and $q_{c,\text{lat}}=30$, so at most 45 raw slots are occupied. \yz{Confirm whether a multi-label sample stored in two partitions is stored twice; this affects the memory numbers below.}
One raw sample contains $500\times500\times10+92\approx2.50\times10^{6}$ values, whereas the whole latent buffer contains $150\times64=9{,}600$ values, i.e., 0.38\% of one raw sample.
At equal numerical precision, the configured capacity of CLAR is therefore about 2.0\% smaller than that of the baselines (about 490\,MB compared with 500\,MB in 32-bit floating point) and provides about four times as many slots.
Since a multi-label sample can occupy several partitions, the number of distinct stored samples can be smaller than the number of slots.
Latent representations are retrieved for every mini-batch with $b=8$, which avoids re-encoding large raw tensors at each step.

\subsection{Baselines}
All baselines use the diagnosis model of Section~\ref{sec:model}, i.e., the same multimodal feature extractor and attention-based classifier, so that the comparison isolates the replay design of Sections~\ref{sec:memory} and~\ref{sec:als}.
No Replay trains on each task sequentially without a buffer and serves as the reference for forgetting. \yz{Strongly recommended: add the ablation CLAR with $\tilde{\sigma}\equiv1$, keeping the dual buffer, teacher, training, and budgets identical, as a row in Table~\ref{tab:results_summary}. None of the current baselines isolates the ALS gate, and this ablation is the empirical counterpart of Corollary~\ref{cor:qstar}. If possible, also estimate $\bar{h}_c$, $V_c$, $\tilde{V}_c$, $\boldsymbol{\delta}_c$, and $q_c^{\star}$ at checkpoints.}
Experience Replay (ER) \cite{chaudhry2019continual} updates a raw buffer by reservoir sampling and retrieves raw samples for every mini-batch.
Class-Balancing Reservoir Sampling (CBRS) \cite{chrysakis2020online} applies reservoir sampling within each class, and can be viewed as ALS-CBRS with $\tilde{\sigma}(\operatorname{ALS}_c)\equiv1$ and a raw buffer only.
iCaRL \cite{rebuffi2017icarl} selects exemplars by herding in the feature space and appends them to the training set of each task, and FBCR \cite{yan2024few} combines virtual prototypes with class-center-based exemplar selection.
The baselines thus differ from CLAR in how samples are selected (random, class-balanced random, or by feature-space centers, compared with ALS-guided selection in CLAR) and in how they are stored (raw samples, compared with the latent-dominant buffer of CLAR).

\subsection{Evaluation Metrics}
After training on task $t$, the model is evaluated on the test sets of all tasks seen so far.
Since normal samples dominate the data, accuracy is uninformative, and we use the weighted F1 score.
Let $S_{k}^{(c)}$ be the number of occurrences of class $c$ in the test set of task $k$, and let $F1_{t,k}^{(c)}$ be the F1 score of class $c$ for the model trained up to task $t$ and evaluated on task $k$, with class-wise predictions obtained by thresholding $\sigma(s_c(z_i))$ under the same protocol for all methods. \yz{State the threshold: fixed at 0.5, or tuned (on which data, per class or global)?}
The weighted F1 score is
\begin{equation}
M_{t,k}=\frac{1}{\sum_{c\in\mathcal{C}_t}S_{k}^{(c)}}\sum_{c\in\mathcal{C}_t}S_{k}^{(c)}F1_{t,k}^{(c)}.
\end{equation}
We report the average weighted F1 (AvgF1), forgetting ($F$), and backward transfer (BWT) after the final task $T$ \cite{lopez2017gradient, chaudhry2018riemannian},
\begin{equation}
\text{AvgF1}=\frac{1}{T+1}\sum_{k=0}^{T}M_{T,k},
\end{equation}
\begin{equation}
F=\frac{1}{T}\sum_{k=0}^{T-1}\Big(\max_{l\in\{k,\dots,T-1\}}M_{l,k}-M_{T,k}\Big),
\end{equation}
\begin{equation}
\text{BWT}=\frac{1}{T}\sum_{k=0}^{T-1}\big(M_{T,k}-M_{k,k}\big).
\end{equation}
AvgF1 measures the final diagnosis performance over all tasks, $F$ measures the largest drop from the best past performance, and BWT measures the net change of earlier tasks between learning them and the end of the stream.
All results are the mean and standard deviation over 5 independent runs. \yz{State what differs across the 5 runs: only the random seed, or also the train/test split, the task partition, and the sample order?}

\begin{table*}[htbp]
\centering
\caption{Performance with $\rho_{\text{base}}=10$ and $\rho_{\text{inc}}\in\{10,15\}$ (mean $\pm$ std over 5 runs). Buffer capacity: 50 raw samples for the baselines; 49 raw samples ($\mathcal{B}_x$) and 150 latent representations ($\mathcal{B}_z$) for CLAR. The best value in each column is in bold.}
\label{tab:results_summary}
\resizebox{\textwidth}{!}{%
\begin{tabular}{l|ccc|ccc}
\hline
& \multicolumn{3}{c|}{\textbf{$\rho_{\text{inc}}=10$}} & \multicolumn{3}{c}{\textbf{$\rho_{\text{inc}}=15$}} \\ \cline{2-7}
\textbf{Method} & \textbf{AvgF1 $\uparrow$} & \textbf{Forgetting $\downarrow$} & \textbf{BWT} & \textbf{AvgF1 $\uparrow$} & \textbf{Forgetting $\downarrow$} & \textbf{BWT} \\ \hline
No Replay & $0.8932 \pm 0.0194$ & $0.0071 \pm 0.0038$ & $0.0468 \pm 0.0118$ & $0.8855 \pm 0.0183$ & $0.0131 \pm 0.0129$ & $0.0382 \pm 0.0186$ \\ \hline
iCaRL & $0.8871 \pm 0.0144$ & $\mathbf{0.0058 \pm 0.0026}$ & $0.0428 \pm 0.0062$ & $0.8891 \pm 0.0060$ & $\mathbf{0.0062 \pm 0.0042}$ & $0.0426 \pm 0.0096$ \\
ER & $0.9069 \pm 0.0090$ & $0.0191 \pm 0.0125$ & $0.0349 \pm 0.0159$ & $0.8937 \pm 0.0200$ & $0.0306 \pm 0.0102$ & $0.0254 \pm 0.0153$ \\
CBRS & $0.8831 \pm 0.0316$ & $0.0394 \pm 0.0350$ & $0.0145 \pm 0.0318$ & $0.8918 \pm 0.0190$ & $0.0269 \pm 0.0096$ & $0.0330 \pm 0.0190$ \\
FBCR & $0.9007 \pm 0.0214$ & $0.0152 \pm 0.0129$ & $0.0490 \pm 0.0226$ & $0.8926 \pm 0.0230$ & $0.0199 \pm 0.0125$ & $0.0394 \pm 0.0227$ \\
\textbf{CLAR} & $\mathbf{0.9248 \pm 0.0174}$ & $0.0120 \pm 0.0096$ & $0.0073 \pm 0.0111$ & $\mathbf{0.9338 \pm 0.0130}$ & $0.0092 \pm 0.0070$ & $0.0161 \pm 0.0093$ \\ \hline
\end{tabular}%
}
\end{table*}

\begin{figure*}[htbp]
    \centering
    \includegraphics[width=0.48\textwidth]{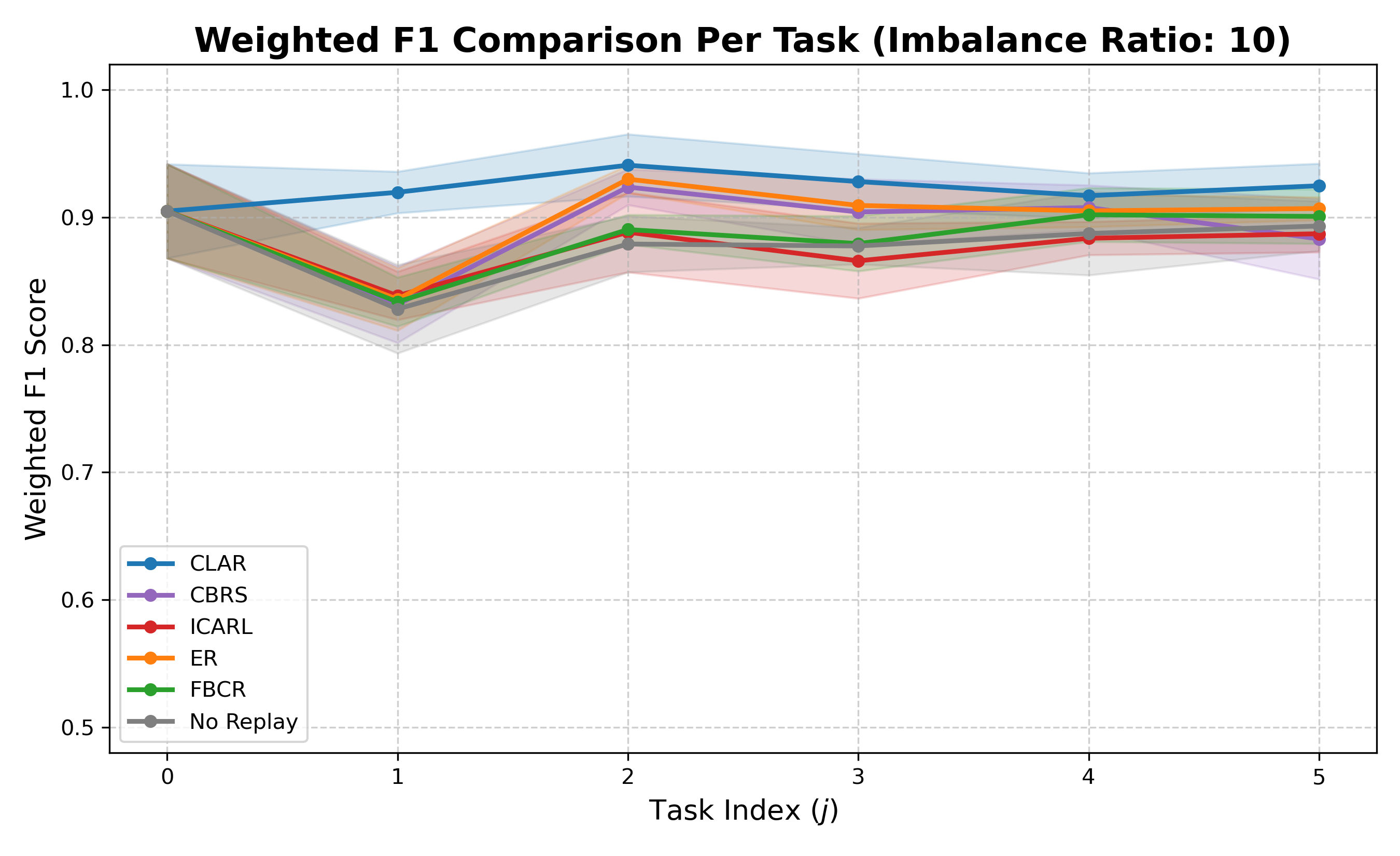}%
    \hfil
    \includegraphics[width=0.48\textwidth]{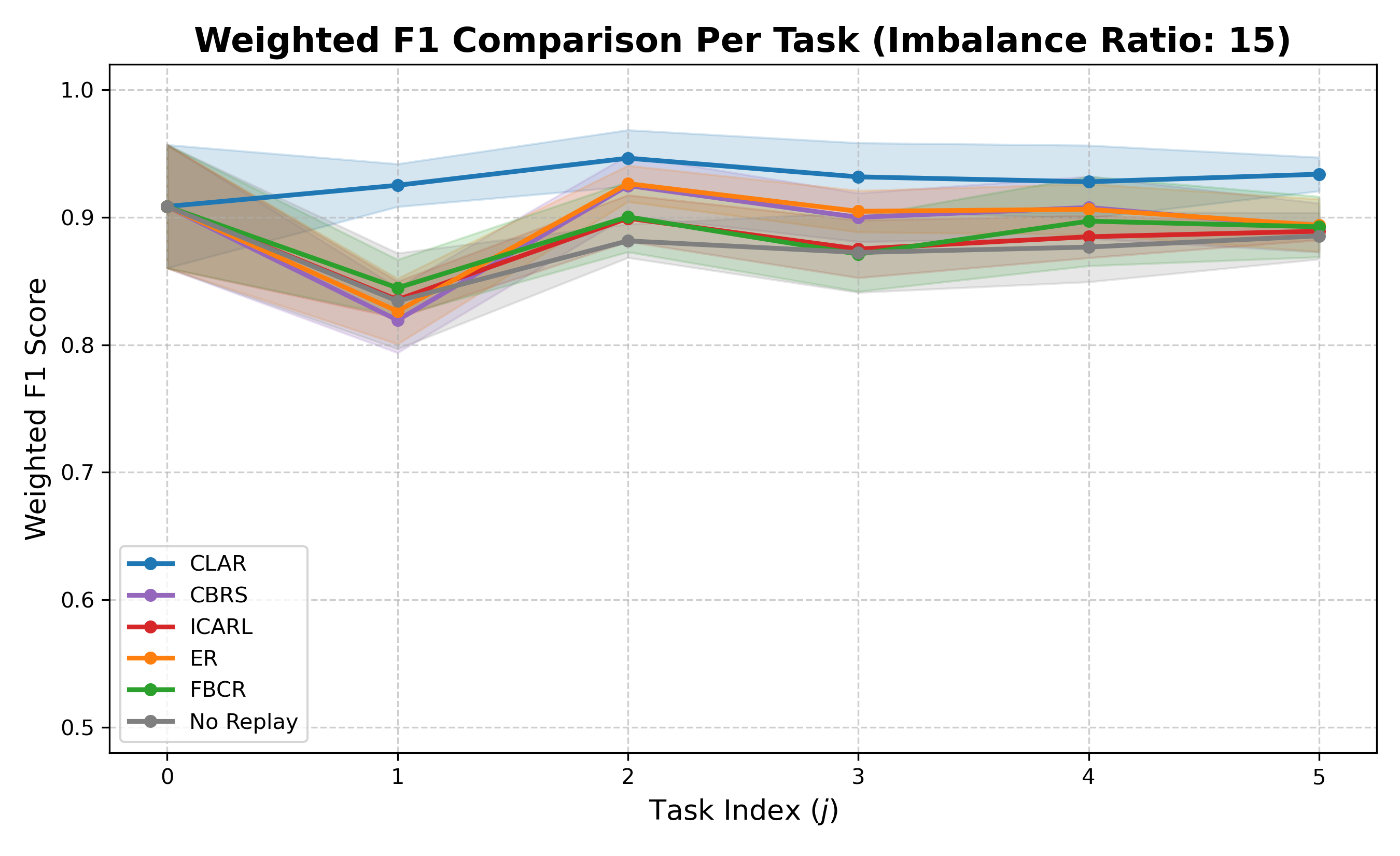}%
    \caption{Weighted F1 score per task for $\rho_{\text{inc}}=10$ (left) and $\rho_{\text{inc}}=15$ (right).}
    \label{fig:weighted_f1_comparison}
\end{figure*}

\subsection{Results}
Table~\ref{tab:results_summary} and Fig.~\ref{fig:weighted_f1_comparison} summarize the results.

\textit{Diagnosis performance.}
CLAR achieves the highest mean AvgF1 under both imbalance levels.
For $\rho_{\text{inc}}=10$, CLAR reaches $0.9248\pm0.0174$, which exceeds the strongest baseline, ER ($0.9069\pm0.0090$), by $0.0179$. 
For $\rho_{\text{inc}}=15$, CLAR reaches $0.9338\pm0.0130$, which exceeds the strongest baseline, again ER ($0.8937\pm0.0200$), by $0.0401$.
The advantage of CLAR thus grows with the imbalance of the incremental tasks.
From $\rho_{\text{inc}}=10$ to $15$, the AvgF1 of CLAR changes by $+0.0090$, while the changes of the baselines range from $-0.0132$ (ER) to $+0.0087$ (CBRS), so CLAR remains stable over the tested range of imbalance.
Since the base-task imbalance is fixed and only two levels are tested, this result supports robustness within the tested conditions.
The raw-only replay baselines remain close to No Replay ($0.8932$ and $0.8855$), and iCaRL and CBRS fall slightly below it for $\rho_{\text{inc}}=10$, which indicates that a few raw exemplars per class bring limited benefit when hazard classes are scarce and multi-label; the latent-dominant buffer of CLAR addresses exactly this regime.

\textit{Forgetting.}
iCaRL attains the lowest forgetting in both settings ($0.0058$ and $0.0062$), and No Replay also forgets little for $\rho_{\text{inc}}=10$ ($0.0071$).
Both, however, have low AvgF1, so their low forgetting reflects limited learning of the incremental classes.
CLAR has the lowest forgetting among the methods with AvgF1 above $0.90$ ($0.0120$ and $0.0092$), and its forgetting decreases as the imbalance grows, whereas the forgetting of ER increases from $0.0191$ to $0.0306$.
Among the methods with AvgF1 above $0.90$, CLAR thus combines the highest AvgF1 with the lowest forgetting.

\textit{Backward transfer.}
All methods, including No Replay, show positive BWT, i.e., the performance on earlier tasks improves on average after later tasks.
This follows from the protocol, in which most incremental tasks revisit classes that have already been seen and thus provide additional training samples for earlier classes.
CLAR shows the smallest BWT in both settings ($0.0073$ and $0.0161$).
Combined with its highest AvgF1, this is consistent with CLAR reaching high performance on each task when it is first learned and changing it little afterward, whereas the baselines start lower on each task and gain more from revisited classes.
BWT is therefore read together with AvgF1 and forgetting.

\textit{Overall assessment.}
CLAR provides the highest mean diagnosis performance under both imbalance levels with competitive forgetting and the smallest net backward change, with a smaller configured buffer than the raw-replay baselines.
iCaRL achieves the lowest forgetting at a lower AvgF1, which shows that the preferred method depends on the metric of interest.
These results support the CLAR framework as a whole under the evaluated memory and imbalance constraints; the contribution of the ALS gate alone is isolated by an ablation with $\tilde{\sigma}\equiv1$, as suggested by Corollary~\ref{cor:qstar}. \yz{Update this sentence with the ablation result once it is available.}

\subsection{Discussions}
The empirical evaluation of the proposed framework provides deep insights into its continual learning capabilities, parameter dynamics, and representational stability under severe class imbalance. 
\begin{figure}[htbp]
    \centering
    \includegraphics[width=\columnwidth]{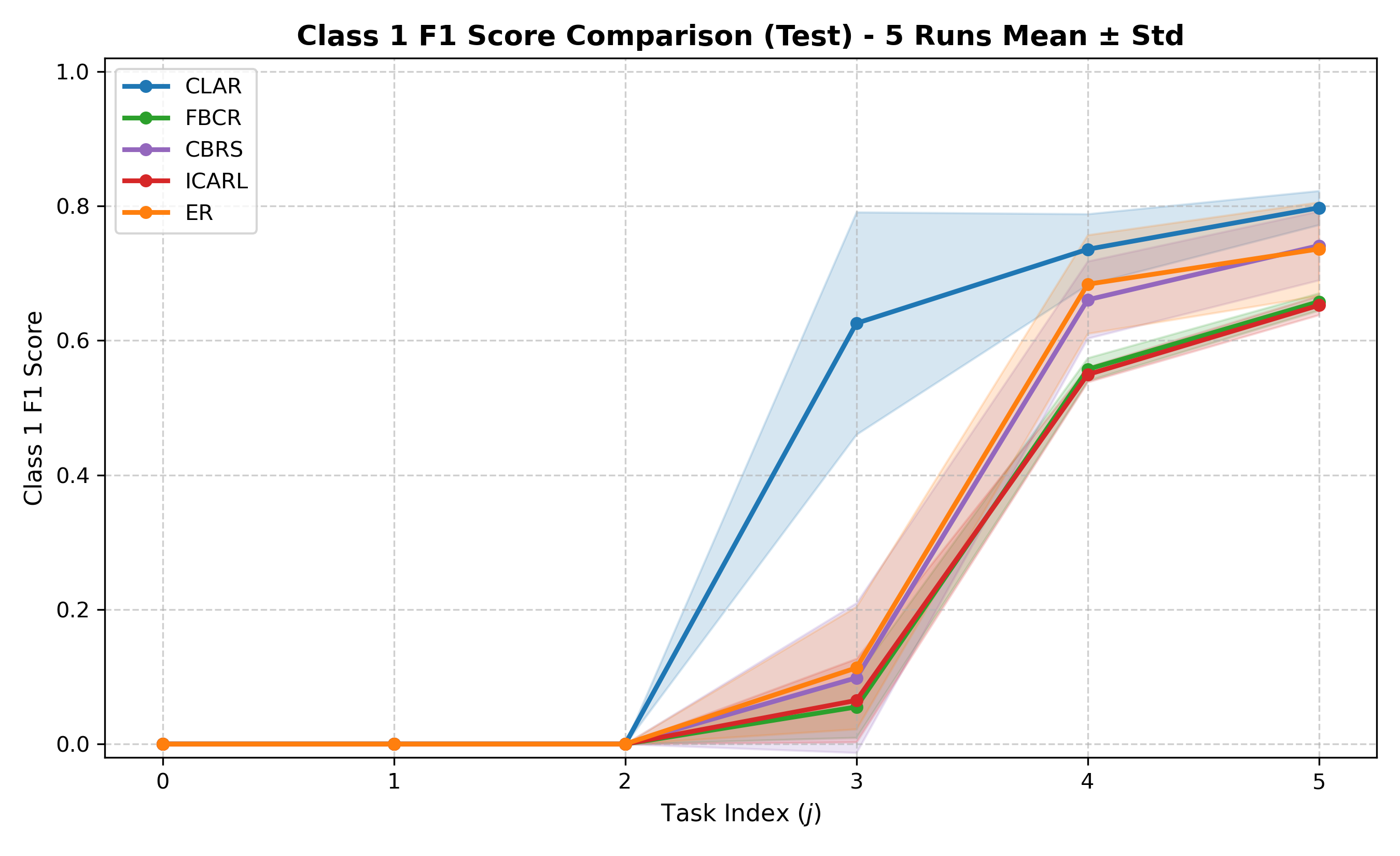}
    \caption{Class 1 per-class F1 Score Comparison across 5 benchmark runs, showing the mean $\pm$ standard deviation for imbalance ratio of 10. Class 1 is introduced as a new class at task index $j = 3$.}
    \label{fig:class1_f1_comparison}
\end{figure}
Figure~\ref{fig:class1_f1_comparison} illustrates the per-class F1 score comparison for Class 1, which is a new class introduced in the incremental learning process, specifically at Task 3, across five benchmark runs under an imbalance ratio of $10$. As observed, CLAR significantly outperforms other continual learning baselines, including ER, iCARL, CBRS, and FBCR, achieving a much higher F1 score and maintaining robust performance. CLAR's dual-replay mechanism and attention formulation enable rapid and stable acquisition of Class 1 without degrading prior knowledge.

\begin{figure*}[htbp]
    \centering
    \includegraphics[width=\textwidth]{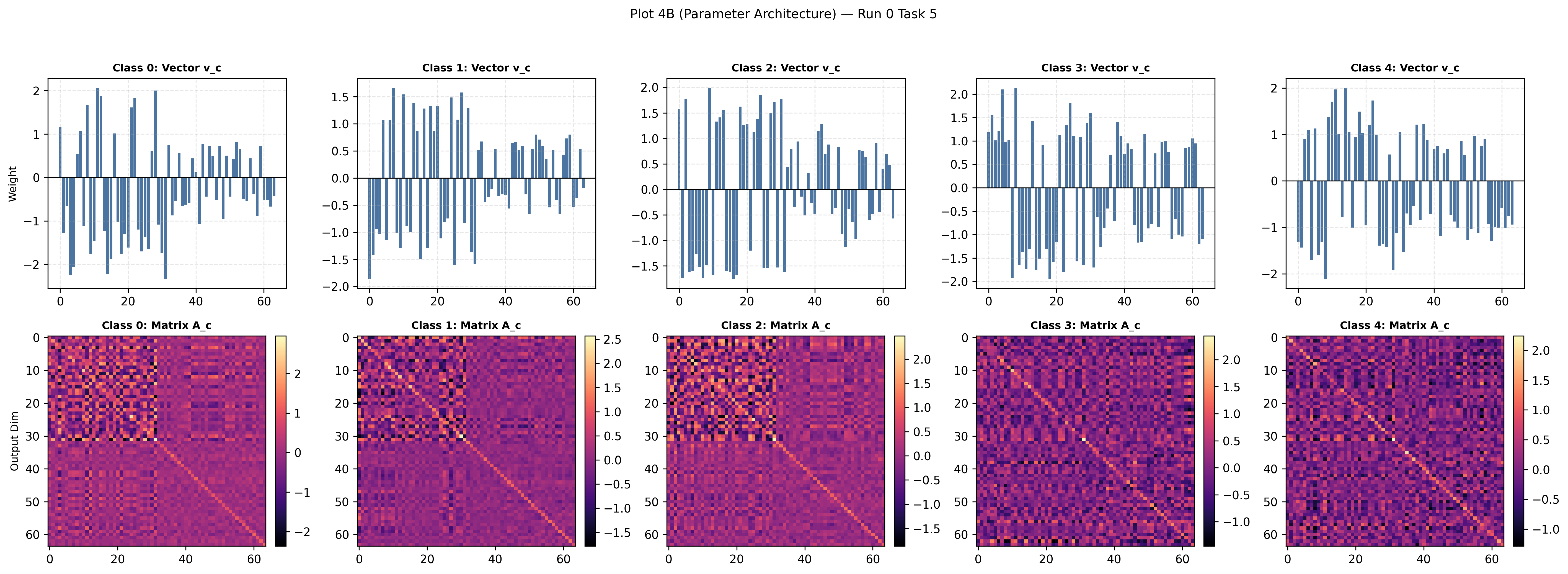}
    \caption{Plot of parameters of the classifier illustrating the learned class-specific attention parameters for after the final task across classes $0$ through $4$. The top row displays the trainable per-class attention vectors $\mathbf{v}_c \in \mathbb{R}^d$ (visualized as bar plots across feature dimensions), which scale feature activations independently. The bottom row shows heatmaps of the per-class attention matrices $\mathbf{A}_c \in \mathbb{R}^{d \times d}$, capturing complex feature interaction geometries and cross-dimensional weightings for each hazard class.}
    \label{fig:parameter_architecture}
\end{figure*}
To understand the internal mechanisms of the attention based classifier, Figure~\ref{fig:parameter_architecture} visualizes the learned class-specific attention parameters after the final task across hazard classes $0$ through $4$. The trainable per-class attention vectors $\mathbf{v}_c$ (top row) isolate critical feature dimensions, while the per-class interaction matrices $\mathbf{A}_c$ (bottom row) capture complex cross-dimensional weightings and feature geometries. This dual formulation allows the model to compute class-aware logits via $s_c(z_i) = (\mathbf{A}_c z_i)^\top (\operatorname{diag}(\mathbf{v}_c) z_i)$, tailoring the decision space specifically to the unique feature correlations of each hazard class.

\begin{figure*}[htbp]
    \centering
    \includegraphics[width=\textwidth]{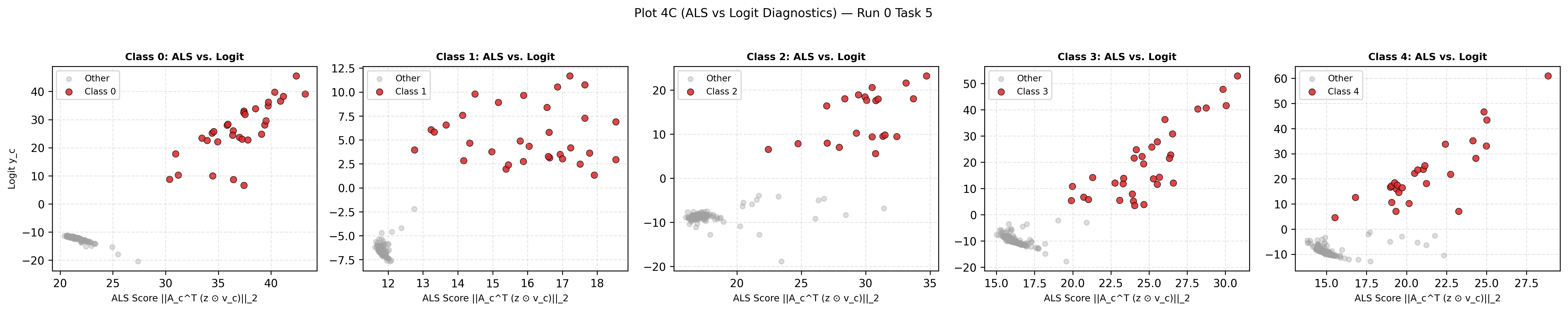}
    \caption{Plot of ALS vs. Logit diagnostics. Gray points represent hazard-free samples, while red points denote hazardous samples for class $c$. The decision boundary is established at $\text{logit} = 0$ (corresponding to a classification threshold of $0.5$)}
    \label{fig:als_vs_logit}
\end{figure*}
Further diagnostic validation is provided in Figure~\ref{fig:als_vs_logit}, which plots the Attention Logit Score (ALS) against the final logit values for all classes after the final task. The clear separation between hazard-free samples (gray points) and hazardous samples (red points) relative to the decision threshold ($\text{logit} = 0$, corresponding to a $0.5$ probability) confirms that the attention-based scoring function reliably discriminates hazardous instances.

\begin{figure*}[htbp]
    \centering
    \includegraphics[width=\textwidth]{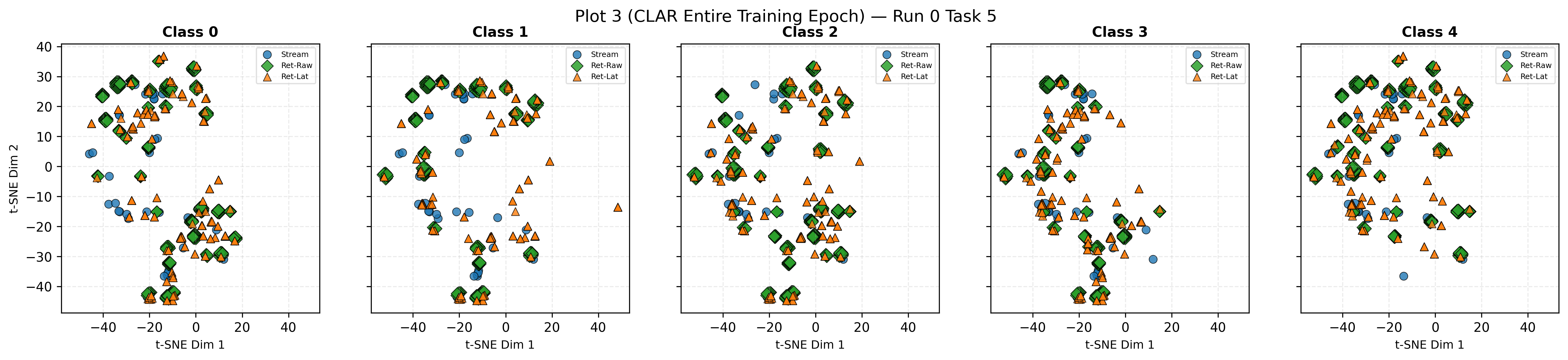}
    \caption{Plot of Training samples used to train the incremental hazard diagnosis model. Illustrating representational consistency during incremental training (Algorithm~\ref{alg:train}). Current streaming training samples are marked in blue circles (Stream), batch-wise retrieved raw samples from $\mathcal{B}_x$ are shown as green diamonds, and retrieved latent representations from $\mathcal{B}_z$ are shown as orange triangles. The spatial overlap between the modalities highlights the consistency and alignment maintained by the dual-replay memory buffer across classes.}
    \label{fig:clar_training_epoch}
\end{figure*}
Finally, Figure~\ref{fig:clar_training_epoch} examines the representational consistency during incremental training across all hazard classes. By visualizing current streaming samples (blue circles), batch-wise retrieved raw samples from $\mathcal{B}_x$ (green diamonds), and retrieved latent representations from $\mathcal{B}_z$ (orange triangles), we observe a strong spatial overlap across modalities. This alignment highlights the effectiveness of the dual-replay memory buffer in preserving feature distribution consistency, preventing representation drift, and ensuring stable continual learning throughout the entire training lifecycle.

\section{Conclusion}\label{sec:conclusion}
This work proposed Class-aware Latent Attention Replay (CLAR), a class-incremental hazard diagnosis framework that supports resilient AI systems in the MII.
CLAR addresses newly emerging and recurring hazard classes, high-dimensional multimodal samples, severe class imbalance, multi-label co-occurrence, and limited memory through three coupled components: a latent-dominant dual replay buffer, ALS-guided class-balanced sampling, and a teacher-anchored latent space.
The ALS links sample selection to the classifier through an upper bound on the logit magnitude.
For an idealized model of the buffer update, the analysis showed that ALS-CBRS keeps a class-balanced buffer that follows an ALS-weighted class distribution, and that it has a lower replay-gradient error and a smaller upper bound on the expected squared margin deviation than class-balanced uniform replay whenever the per-class capacity lies below a critical value determined by the gate and the gradient statistics of the class.

On hazard data from an MII testbed with $\rho_{\text{base}}=10$ and $\rho_{\text{inc}}\in\{10,15\}$, CLAR achieved the highest mean AvgF1 ($0.9248$ and $0.9338$) among all methods with a configured buffer about 2\% smaller than those of the raw-replay baselines.
Its forgetting ($0.0120$ and $0.0092$) was the lowest among methods with comparable diagnosis performance, while iCaRL forgot less at a lower AvgF1.
All methods showed positive BWT, and CLAR showed the smallest one, reflecting high initial performance on each task with little subsequent change.
These results highlight the value of evaluating incremental diagnosis with complementary metrics.
Future work will estimate these gradient statistics at training checkpoints to verify the critical capacity empirically, examine the separation of class clusters in the latent space, and extend CLAR to cyber-physical layer hazards and to the diagnosis of hazard severity levels.
\bibliographystyle{unsrt}
\bibliography{ref}
\appendices

\section*{Appendix}

\section{Proof of Proposition~\ref{prop:balance}}
The quotas never increase, because $|\mathcal{C}_t|$ is nondecreasing in $t$.
A partition of a new class starts empty with $N_c=0$.
An arrival of class $c$ increases $|\mathcal{B}[c]|$ by one if $|\mathcal{B}[c]|<q_c$ and leaves it unchanged otherwise, so with a fixed quota, $|\mathcal{B}[c]|=\min\{q_c,N_c\}$ by induction over arrivals.
When the quota decreases from $q_c$ to $q_c'<q_c$, eviction leaves $\min\{q_c',\min\{q_c,N_c\}\}=\min\{q_c',N_c\}$ entries, and the induction continues with $q_c'$. \hfill$\square$

\section{Proof of Lemma~\ref{lem:law} and Exact Replay Error}
For a measurable set $\mathcal{A}$ and $N_c>q_c$, conditioning on admission gives
\[
\Pr\big((z,y)\in\mathcal{A}\mid\text{admission}\big)=\frac{\int_{\mathcal{A}}(q_c/N_c)\,h_c\,d\pi_c}{q_c\bar{h}_c/N_c}=\tilde{\pi}_c(\mathcal{A}),
\]
which is the acceptance--rejection identity \cite{vonneumann1951various}.
The process can therefore be represented by independent replacement events with probabilities $q_c\bar{h}_c/N_c$, independent uniform slot choices, and independent replacement samples with law $\tilde{\pi}_c$.
A fixed slot is replaced at arrival $j$ with probability $\bar{h}_c/j$, which gives Eq.~\eqref{eq:retention}; hence the number $\nu_c$ of surviving initial entries satisfies $\mathbb{E}\nu_c=q_c\kappa_c(N_c)$.
Couple $\mathcal{B}[c]$ with a partition initialized with i.i.d.\ entries from $\tilde{\pi}_c$ and subjected to the same replacements.
The coupled partition keeps the law $\tilde{\pi}_c^{\otimes q_c}$, and the two partitions coincide once every initial entry has been replaced, so the total variation distance is at most $\Pr(\nu_c\ge1)\le\mathbb{E}\nu_c=q_c\kappa_c(N_c)$.
Since $\sum_j\bar{h}_c/j$ diverges, $\kappa_c(N_c)\to0$, which proves convergence. \hfill$\square$

\begin{lemma}[Exact replay error]\label{lem:exact}
Let $\mathbf{f}$ be a statistic with $\mathbb{E}_{\pi_c}\|\mathbf{f}\|_2^2<\infty$, with means $\mathbf{m}$ and $\tilde{\mathbf{m}}$ and covariance traces $V_{\mathbf{f}}$ and $\tilde{V}_{\mathbf{f}}$ under $\pi_c$ and $\tilde{\pi}_c$, and let $\boldsymbol{\delta}_{\mathbf{f}}=\tilde{\mathbf{m}}-\mathbf{m}$ and $\bar{\mathbf{f}}=q_c^{-1}\sum_{\mathcal{B}[c]}\mathbf{f}$. Then
\begin{equation}
\mathbb{E}\|\bar{\mathbf{f}}-\mathbf{m}\|_2^2=\frac{(1-\kappa_c)\tilde{V}_{\mathbf{f}}+\kappa_c V_{\mathbf{f}}}{q_c}+\beta_c\|\boldsymbol{\delta}_{\mathbf{f}}\|_2^2,
\label{eq:exact}
\end{equation}
where $\kappa_c=\kappa_c(N_c)$, $\beta_c=\beta_c(N_c)=\mathbb{E}(1-\nu_c/q_c)^2$, $(1-\kappa_c)/q_c\le\beta_c\le1-\kappa_c$, and $\beta_c\to1$ as $N_c\to\infty$.
\end{lemma}
\begin{proof}
For slot $j$, let $\mathbf{F}_j=\mathbf{f}(\text{slot }j)-\tilde{\mathbf{m}}$. Conditional on the replacement schedule, the slots are independent, with law $\pi_c$ for surviving initial entries and $\tilde{\pi}_c$ otherwise. Hence $\mathbb{E}\|\mathbf{F}_j\|_2^2=\kappa_c(V_{\mathbf{f}}+\|\boldsymbol{\delta}_{\mathbf{f}}\|_2^2)+(1-\kappa_c)\tilde{V}_{\mathbf{f}}$.
For $q_c\ge2$, two distinct initial entries both survive with probability $\kappa_c^{(2)}=\prod_{j=q_c+1}^{N_c}(1-2\bar{h}_c/j)$, so $\mathbb{E}\mathbf{F}_j^{\top}\mathbf{F}_k=\kappa_c^{(2)}\|\boldsymbol{\delta}_{\mathbf{f}}\|_2^2$ for $j\neq k$, and $\mathbb{E}\mathbf{F}_j=-\kappa_c\boldsymbol{\delta}_{\mathbf{f}}$.
Expanding the second moment of $q_c^{-1}\sum_j\mathbf{F}_j+\boldsymbol{\delta}_{\mathbf{f}}$ gives Eq.~\eqref{eq:exact} with $\beta_c=1-2\kappa_c+\kappa_c/q_c+(q_c-1)\kappa_c^{(2)}/q_c$, where the last term vanishes for $q_c=1$.
The identities $\mathbb{E}\nu_c=q_c\kappa_c$ and $\mathbb{E}\nu_c^2=q_c\kappa_c+q_c(q_c-1)\kappa_c^{(2)}$ identify $\beta_c$ with $\mathbb{E}(1-\nu_c/q_c)^2$.
Since $1-\nu_c/q_c\in\{0,1/q_c,\dots,1\}$, its square lies between $(1-\nu_c/q_c)/q_c$ and $1-\nu_c/q_c$, which gives the bounds. Finally, $\kappa_c\to0$ implies $\nu_c/q_c\to0$ in probability, and boundedness gives $\beta_c\to1$.
\end{proof}

\section{Proof of Theorem~\ref{thm:main}}
In Eq.~\eqref{eq:grad}, the residual factor is bounded by $\max\{w_{c'},1\}$, and $\|\psi_{c'}(z)\|_2^2\le\|z\|_2^2\|\mathbf{A}_{c'}z\|_2^2\le\|\mathbf{A}_{c'}\|_{\mathrm{op}}^2\|z\|_2^4$.
Hence $\mathbf{g}$ is square integrable under $\pi_c$ and, since $h_c\le1$, under $\tilde{\pi}_c$.
Lemma~\ref{lem:exact} with $\mathbf{f}=\mathbf{g}$, $\kappa_c\to0$, and $\beta_c\to1$ gives the limit in Eq.~\eqref{eq:limit}. For CBRS, $\tilde{\pi}_c=\pi_c$ and $\boldsymbol{\delta}_c=\mathbf{0}$, so Eq.~\eqref{eq:exact} equals $V_c/q_c$ for every $N_c\ge q_c$.

For the margin, let $\boldsymbol{\Delta}=\hat{\mathbf{v}}-\mathbf{v}^{\star}$. The curvature condition and $\nabla\hat{\mathcal{J}}(\hat{\mathbf{v}})=\mathbf{0}$ give
\[
\mu\|\boldsymbol{\Delta}\|_2^2\le\boldsymbol{\Delta}^{\top}\big(\nabla\hat{\mathcal{J}}(\hat{\mathbf{v}})-\nabla\hat{\mathcal{J}}(\mathbf{v}^{\star})\big)\le\|\boldsymbol{\Delta}\|_2\,\|\nabla\hat{\mathcal{J}}(\mathbf{v}^{\star})\|_2 .
\]
Since $\nabla\mathcal{J}(\mathbf{v}^{\star})=\mathbf{0}$, the common term $\Omega$ cancels and $\nabla\hat{\mathcal{J}}(\mathbf{v}^{\star})=\bar{\mathbf{g}}-\mathbb{E}_{\pi_c}\mathbf{g}$, so $\mathbb{E}\|\boldsymbol{\Delta}\|_2^2\le\mathcal{E}_c(N_c)/\mu^2$.
Finally, $\gamma(\hat{\mathbf{v}})-\gamma(\mathbf{v}^{\star})=(2y_{*,c'}-1)\psi_{c'}(z_*)^{\top}\boldsymbol{\Delta}$, and the Cauchy--Schwarz inequality gives Eq.~\eqref{eq:margin}. The same argument applies to CBRS. \hfill$\square$

\section{Proof of Corollary~\ref{cor:qstar}}\label{app:cor}
Lemma~\ref{lem:exact} with $\mathbf{f}=\mathbf{g}$ gives
\begin{equation}
\mathcal{E}_c(N_c)-\mathcal{E}_c^{\mathrm{CBRS}}(N_c)=\frac{1-\kappa_c}{q_c}\big(\tilde{V}_c-V_c\big)+\beta_c\|\boldsymbol{\delta}_c\|_2^2 .
\label{eq:diff}
\end{equation}
If $q_c<q_c^{\star}$, then $(V_c-\tilde{V}_c)/q_c>\|\boldsymbol{\delta}_c\|_2^2$, and $\beta_c\le1-\kappa_c$ yields
\[
\mathcal{E}_c(N_c)-\mathcal{E}_c^{\mathrm{CBRS}}(N_c)\le(1-\kappa_c)\Big(\|\boldsymbol{\delta}_c\|_2^2-\frac{V_c-\tilde{V}_c}{q_c}\Big)<0
\]
for $N_c>q_c$, since $\kappa_c<1$ there. The margin bounds inherit this ordering whenever $\psi_{c'}(z_*)\neq\mathbf{0}$.
As $N_c\to\infty$, the difference tends to $\|\boldsymbol{\delta}_c\|_2^2-(V_c-\tilde{V}_c)/q_c$, which is positive for $q_c>q_c^{\star}$.
For the second form of $q_c^{\star}$, a change of measure gives
\[
\begin{aligned}
\tilde{V}_c&=\mathbb{E}_{\tilde{\pi}_c}\|\mathbf{g}-\mathbb{E}_{\pi_c}\mathbf{g}\|_2^2-\|\boldsymbol{\delta}_c\|_2^2\\
&=V_c+\frac{\operatorname{Cov}_{\pi_c}(h_c,\|\mathbf{g}-\mathbb{E}_{\pi_c}\mathbf{g}\|_2^2)}{\bar{h}_c}-\|\boldsymbol{\delta}_c\|_2^2 ,
\end{aligned}
\]
and substitution into Eq.~\eqref{eq:qstar} completes the proof. At a finite $N_c>q_c$, Eq.~\eqref{eq:diff} gives the exact criterion $\mathcal{E}_c(N_c)<\mathcal{E}_c^{\mathrm{CBRS}}(N_c)$ if and only if $(1-\kappa_c)(V_c-\tilde{V}_c)>q_c\beta_c\|\boldsymbol{\delta}_c\|_2^2$. \hfill$\square$

\end{document}